\documentclass[10pt,twocolumn,letterpaper]{article}

\usepackage[pagenumbers]{cvpr} %

\usepackage{microtype}

\usepackage{url}

\usepackage{booktabs}
\usepackage[table]{xcolor}
\usepackage{graphicx}
\usepackage{adjustbox}
\usepackage{array}
\usepackage{pifont}
\usepackage{multirow}
\usepackage{placeins}
\usepackage{tabularx}
\usepackage{mathtools}
\usepackage{dsfont}
\usepackage{svg}
\usepackage{pgfplots}

\pgfplotsset{compat=1.17}

\definecolor{cOthers}{RGB}{0,0,0}
\definecolor{cBarrier}{RGB}{112,128,144}
\definecolor{cBicycle}{RGB}{220,20,60}
\definecolor{cBus}{RGB}{255,127,80}
\definecolor{cCar}{RGB}{255,158,0}
\definecolor{cConsVeh}{RGB}{233,150,70}
\definecolor{cMotorcycle}{RGB}{255,61,99}
\definecolor{cPedestrian}{RGB}{0,0,230}
\definecolor{cTrafficCone}{RGB}{47,79,79}
\definecolor{cTrailer}{RGB}{255,140,0}
\definecolor{cTruck}{RGB}{255,99,71}
\definecolor{cDrivSurf}{RGB}{0,207,191}
\definecolor{cOtherFlat}{RGB}{175,0,75}
\definecolor{cSidewalk}{RGB}{75,0,75}
\definecolor{cTerrain}{RGB}{112,180,60}
\definecolor{cManmade}{RGB}{222,184,135}
\definecolor{cVegetation}{RGB}{0,175,0}

\definecolor{cGO}{RGB}{0,0,0}
\definecolor{cVehicle}{RGB}{255,158,0}
\definecolor{cPedestrianW}{RGB}{0,0,230}
\definecolor{cSign}{RGB}{47,79,79}
\definecolor{cBicyclist}{RGB}{220,20,60}
\definecolor{cTrafficLight}{RGB}{255,69,0}
\definecolor{cPole}{RGB}{255,140,0}
\definecolor{cConsCone}{RGB}{233,150,70}
\definecolor{cBicycleW}{RGB}{255,61,99}
\definecolor{cMotorcycleW}{RGB}{112,128,144}
\definecolor{cBuilding}{RGB}{222,184,135}
\definecolor{cVegetationW}{RGB}{0,175,0}
\definecolor{cTreeTrunk}{RGB}{165,42,42}
\definecolor{cRoad}{RGB}{0,207,191}
\definecolor{cWalkable}{RGB}{75,0,75}

\definecolor{bannergray}{gray}{0.88}

\newcommand{\colhead}[2]{%
  \begin{tabular}[b]{@{}c@{}}
    \rotatebox{90}{\footnotesize #2} \\[2pt] %
    \textcolor{#1}{\rule{2.2mm}{2.2mm}}      %
  \end{tabular}%
}
\newcommand{\cmark}{\text{\ding{51}}}
\newcommand{\xmark}{\text{\ding{55}}}

\newcommand{\approach}{PseudoWorld\xspace}

\definecolor{cvprblue}{rgb}{0.21,0.49,0.74}
\usepackage[pagebackref,breaklinks,colorlinks,allcolors=cvprblue]{hyperref}

\def\confName{CVPR}
\def\confYear{2027}

\title{PseudoWorld: 3D Pseudo-Supervision for Semantic Occupancy World Models}

\author{Nicolas Schischka \quad Rohit Mohan \quad Abhinav Valada\\
University of Freiburg\\
}

\begin{document}
\maketitle

\begin{abstract}
3D semantic occupancy world models provide rich semantic and spatial scene understanding of the environment evolution.
However, their training typically requires expensive human-labeled 3D semantic annotations derived from LiDAR data. Existing approaches that avoid manual 3D annotations rely on 2D supervision, which provides only indirect constraints on 3D geometry and therefore underperforms.
We propose \approach, a novel framework that leverages foundation models to generate 3D semantic pseudo-labels from posed multi-view images, enabling semantic occupancy world model training without human annotations or LiDAR data.  Our pseudo-labeling approach identifies moving objects through long-term motion segmentation and completes their 3D shapes. We employ hybrid-resolution supervision that captures fine object details and harness the per-voxel motion annotations to introduce motion-guided losses, which steer the training of a sparse occupancy world model. \approach achieves state-of-the-art performance for weakly-supervised occupancy prediction on the established Occ3D-nuScenes benchmark and outperforms prior 3D label-free methods for 4D semantic occupancy forecasting on both Occ3D-nuScenes and Occ3D-Waymo, without using LiDAR depth or 2D semantic labels.
Project page:~\href{https://pseudoworld.cs.uni-freiburg.de}{\nolinkurl{pseudoworld.cs.uni-freiburg.de}}.
\end{abstract}    
\section{Introduction}
World models~\cite{ha2018world, lecun2022path, hafner2025mastering, schmidt2026relationally, li2025enhancing} enable physical AI systems to anticipate how their surroundings will evolve.
For autonomous driving and embodied agents, 3D semantic occupancy world models~\cite{zheng2024occworld, gu2024dome, li2025semi, mohan2026forecastocc, du2025sparseworld, dang2026sparseworld} represent scene geometry, semantics, and dynamics in an explicit 3D space, offering more efficient and geometrically consistent representations than generative driving video models~\cite{agarwal2025cosmos, gao2024vista, hassan2025gem, mousakhan2025overcoming, mittal2026orbis}.
Recent advances in weakly- and self-supervised learning~\cite{huang2024selfocc, gan2025gaussianocc, boeder2025gaussianflowocc, lilja2026queryocc, boeder2025shelfocc, abdelsamad2025multi} have reduced the annotation requirements of 3D semantic occupancy prediction models~\cite{irshad2026neural, zhang2023occformer, huang2024gaussianformer, luz2026latent, luz2026streaming, wang2024opus}. However, occupancy world models still typically rely on dense human-annotated 3D semantic labels derived from LiDAR data.
This dependency limits their ability to scale with unlabeled data from fleet vehicles equipped only with cameras.\looseness=-1

\begin{figure}
    \centering
    {\includegraphics[width=\linewidth]{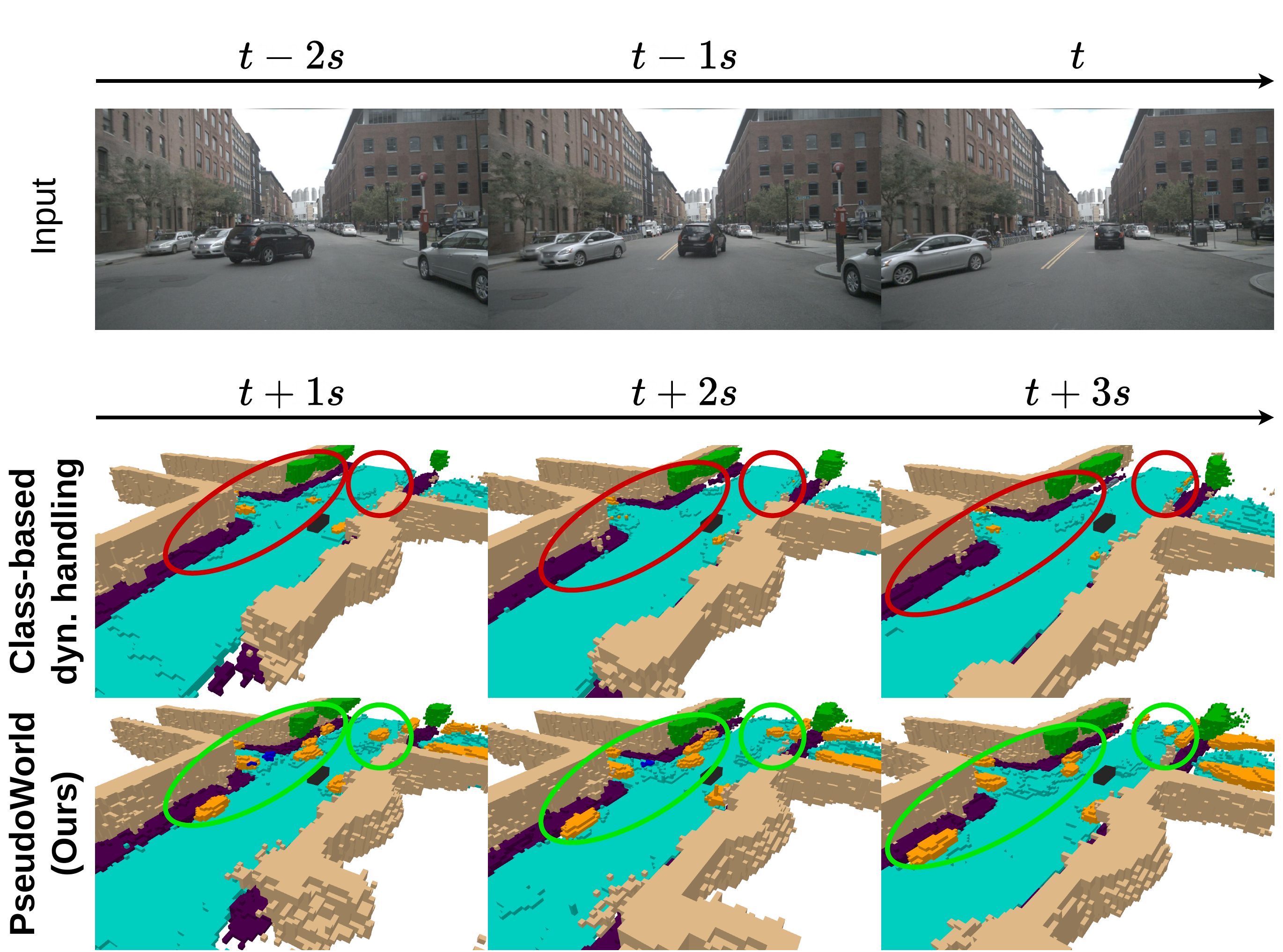}}
    \caption{Previous label-free 3D occupancy prediction methods~\cite{boeder2025shelfocc, caunes2026freeocc} use semantic classes as proxies for motion during pseudo-label generation.
    When directly adopted for 4D semantic occupancy forecasting, the trained world model performs poorly at forecasting potentially dynamic classes.
    \approach explicitly distinguishes stationary from moving objects and uses motion-aware pseudo-labels and supervision to better capture scene dynamics.}
    \label{fig:teaser}
    \vspace{-0.3cm}
\end{figure}

Existing approaches that avoid direct 3D supervision instead supervise the world model~\cite{li2025semi} or its underlying occupancy prediction model~\cite{zheng2024occworld, yan2025renderworld} through 2D rendering.
However, rendering-based supervision constrains 3D geometry only indirectly and can produce artifacts such as depth bleeding, limiting the overall performance of these approaches.
Moreover, prior work~\cite{yan2025renderworld, li2025semi} obtains their 2D semantic supervision by projecting labeled 3D ground truth, retaining the dependence on manual annotations. 
Recent label- and LiDAR-free occupancy prediction methods~\cite{hayes2026easy3d, boeder2025shelfocc, caunes2026freeocc} instead use foundation models for open-vocabulary segmentation and depth estimation to generate pseudo-labeled 3D semantic point clouds. Here, \emph{label-free} denotes supervision derived solely from raw sensor data and off-the-shelf pretrained models, while \emph{LiDAR-free} excludes LiDAR point clouds and LiDAR-derived depth from both inputs and supervision. These methods enable direct 3D supervision and alleviate the limitations of 2D rendering, but remain limited to single-frame predictions.

ShelfOcc~\cite{boeder2025shelfocc} and FreeOcc~\cite{caunes2026freeocc} generate 3D pseudo-labels by accumulating points from static scene regions over time to avoid undesired trails from dynamic objects. However, they use semantic classes as a proxy for motion, treating all potentially moving objects as dynamic.
In real-world scenarios, this assumption does not hold: parked (65.8\%) and stopped (10.3\%) vehicles represent the vast majority of all vehicles in nuScenes. Consequently, these types of vehicles and other stationary but potentially dynamic objects are excluded from temporal accumulation. Their representations therefore remain sparse in the point cloud and disappear from the pseudo-labels once they leave the field of view.
Similarly, occluded parts of moving objects have incomplete shapes as they are not observed.
For world model training, these limitations obscure actual motion and introduce visibility-driven disappearance and reappearance into the training targets, impairing dynamics learning, as illustrated in \Cref{fig:teaser}.
Moreover, moving objects are often small and rare, making them underrepresented and poorly resolved in coarse voxel grids.

To address these limitations, we propose \textbf{\approach}, a framework for label- and LiDAR-free 3D semantic occupancy world model training.
We first perform motion-aware 3D pseudo-label generation, in which we explicitly disentangle static from moving objects.
We achieve this via long-term motion segmentation from bird's-eye-view (BEV) instance tracks, allowing us to accumulate points for static objects without encountering trails from moving objects.
Since this benefits only static objects, we employ moving object completion to recover more complete shapes of dynamic objects, whose sparse point clouds are otherwise derived solely from the current timestamp and viewpoint.
Our motion-aware resolution-adaptive voxelization uses the enhanced point cloud to derive motion annotations per voxel and generate pseudo-labels at different resolutions.
We use a fine-grained voxel grid to capture object classes with more detail and a coarse voxel grid to efficiently represent large stuff classes.
Our world model training subsequently leverages the motion annotations in motion-guided losses and the fine and coarse grids for hybrid-resolution supervision.

Extensive evaluations on Occ3D-nuScenes demonstrate state-of-the-art performance for label-free 3D semantic occupancy prediction, with a relative performance gain of 12.6\% in RayIoU.
For 4D semantic occupancy forecasting, \approach outperforms previous 3D label-free world models by 81.0\%, without using LiDAR depth supervision and 2D ground truth segmentation masks.
To further demonstrate that our pseudo-labeling framework directly transfers to different camera setups and environments, we establish the first label- and LiDAR-free results for 4D semantic occupancy forecasting on Occ3D-Waymo.

Our contributions are summarized as follows:
\begin{itemize}
    \item We propose a motion-aware pseudo-label generation framework for 3D semantic occupancy prediction that disentangles moving from static objects via the long-term BEV displacements of 2D instance tracks and completes the shapes of moving vehicles.
    \item We leverage our framework to create labels at coarse and fine-grained resolutions, with per-voxel motion annotations to guide label-free training of a sparse 3D semantic occupancy world model.
    \item We achieve state-of-the-art performance on Occ3D-nuScenes for label-free 3D semantic occupancy prediction and 4D forecasting across multiple sparse occupancy world models, and report the first label-free 4D semantic occupancy forecasting results on Occ3D-Waymo.
    \item We will publicly release the code, pseudo-labels, and trained models upon acceptance.
\end{itemize}
\section{Related Work}

\begin{figure*}[t]
    \centering
    {\includegraphics[width=0.9\linewidth]{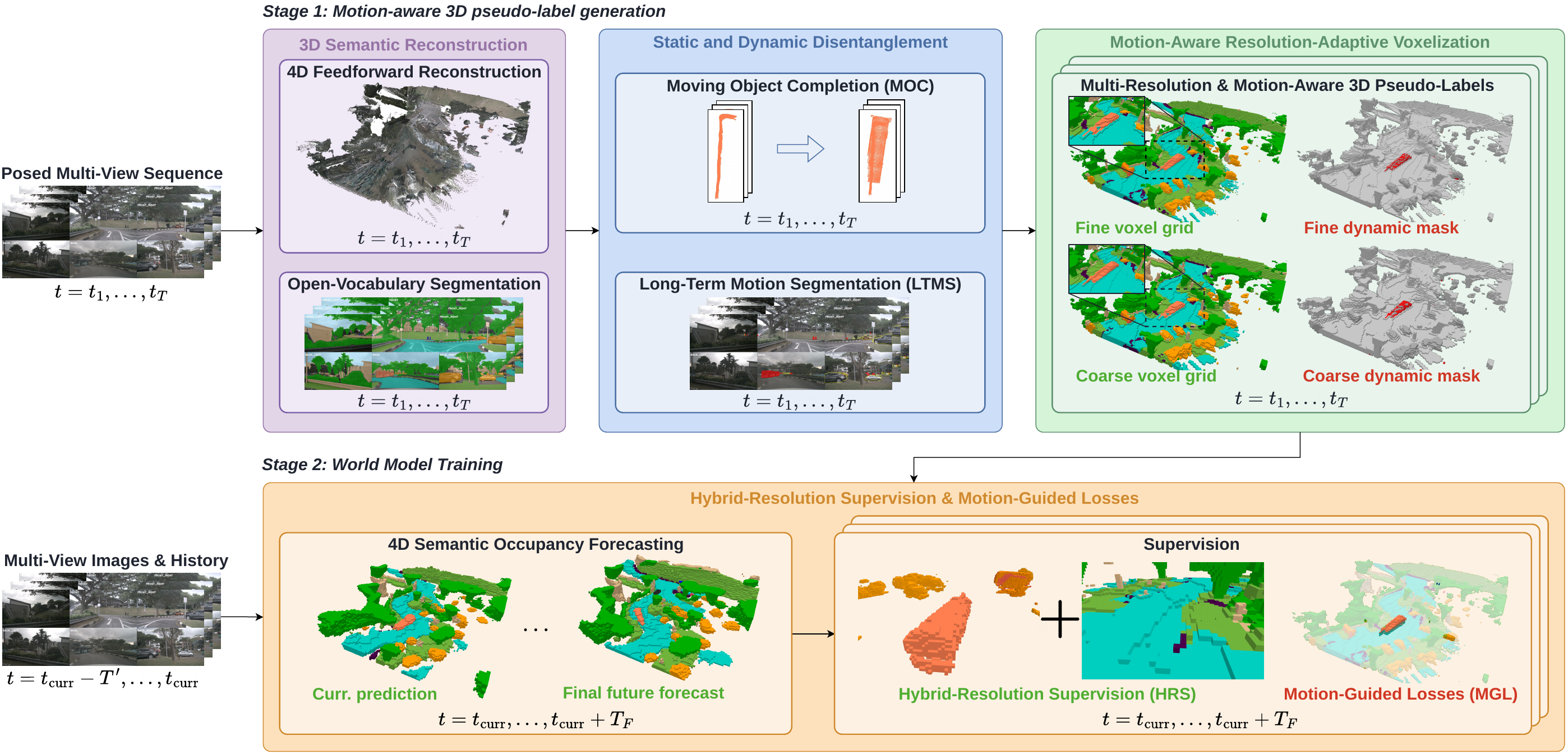}}    \caption{Overview of \textbf{\approach}. We first generate 3D semantic occupancy pseudo-labels (\Cref{subsec:label_creation}) by creating a 3D semantic reconstruction (\Cref{subsubsec:recon}), disentangling static and dynamic scene parts (\Cref{subsubsec:motion_seg}), and voxelizing the scene into multi-resolution semantic voxel grids with motion annotations (\Cref{subsec:voxelization}). Subsequently, we leverage these labels to train a 3D semantic occupancy world model with a hybrid resolution voxel grid and employ motion-guided losses (\Cref{subsec:model_training}).}
    \label{fig:architecture}
    \vspace{-0.2cm}
\end{figure*}

{\noindent\textbf{Label- \& LiDAR-free 3D semantic occupancy} methods aim to reduce the dependency on human-annotated 3D labels and LiDAR data.
Early approaches project the estimated 3D occupancy to 2D through volumetric rendering, enforce photometric consistency, and use weak supervision with open-vocabulary segmentation models~\cite{huang2024selfocc, zhang2025occnerf}.
Instead of volumetric rendering, GaussianOcc~\cite{gan2025gaussianocc} and GaussTR~\cite{jiang2025gausstr} build on 3D Gaussian splatting, while GaussianFlowOcc~\cite{boeder2025gaussianflowocc} additionally models scene dynamics.
To alleviate the negative influence of incorrect pseudo-labels, LetOccVote~\cite{zhang2026letoccvote} proposes depth and semantic voting to identify unreliable regions.
Instead of rendering supervision, QueryOcc~\cite{lilja2026queryocc} supervises occupancy directly with independent 4D spatio-temporal queries.
Another line of research focuses on overcoming information loss from 2D projection by generating 3D pseudo-labels for native supervision in 3D space~\cite{hayes2026easy3d, boeder2025shelfocc, caunes2026freeocc}.
Easy3D-Labels~\cite{hayes2026easy3d} uses monocular depth estimation, while ShelfOcc~\cite{boeder2025shelfocc} employs a feedforward foundation model for depth estimation and accumulates points from static classes over time.
FreeOcc~\cite{caunes2026freeocc} further builds on this by identifying instances and explores training-free panoptic occupancy prediction.
Following a similar motivation, TT-Occ~\cite{zhang2026test} performs training-free semantic occupancy prediction by optimizing 3D Gaussians with foundation models, including optical flow estimation for dynamics modeling.
In this work, we follow this line of research and develop a 3D pseudo-labeling method that explicitly disentangles static and dynamic parts via long-term motion segmentation from 2D instance tracks.
}

{\parskip=2pt
\noindent\textbf{Self-supervised occupancy forecasting and world models} have been developed to estimate the future scene evolution without human annotations, commonly via supervision derived from LiDAR observations.
Early works focus on future geometric occupancy forecasting in bird's-eye view~\cite{khurana2022differentiable} and 3D space~\cite{khurana2023point}, supervised from posed LiDAR sweeps.
After tokenizing past LiDAR observations, CoPilot4D~\cite{zhang2024copilot4d} leverages a diffusion model for point cloud forecasting.
Moving beyond predictions at fixed discrete timestamps, UnO~\cite{agro2024uno} learns a queryable, continuous occupancy field for LiDAR-based occupancy forecasting.
DIO~\cite{diehl2025dio} extends UnO with scene flow estimation and can be prompted for specific instances, while TriO~\cite{sykora2026trio} further performs obstacle segmentation and adds radar and camera data as inputs and for supervision.
Unlike LiDAR-supervised geometric forecasting methods, our approach uses only posed multi-view images during both training and inference to forecast multi-class semantic occupancy.
}

{\parskip=2pt
\noindent\textbf{Camera-based semantic occupancy world models} estimate future occupancy with semantic class labels from RGB images. %
OccWorld~\cite{zheng2024occworld} takes past 3D semantic occupancy grids as input, tokenizes them with a VQ-VAE into BEV, and auto-regressively forecasts future occupancy states.
OccLLaMA~\cite{wei2024occllama} further integrates language and adopts a pre-trained LLM as a generative world model, while RenderWorld~\cite{yan2025renderworld} leverages 2D rendering supervision to train the underlying occupancy network.
Instead of using a discrete tokenizer, DOME~\cite{gu2024dome} employs a continuous latent space in combination with a spatio-temporal diffusion model, which COME~\cite{shi2026come} extends by conditioning the diffusion model on ego-motion-independent scene evolution.
\textit{I$^2$}-World~\cite{liao2025i2} improves efficiency by using an intra- and inter-scene tokenization and applies an encoder-decoder architecture conditioned on ego transformations.
GAST~\cite{chen2026geometry} further introduces explicit-implicit generation and spatio-temporal modeling for improved temporal and spatial consistency.
Another line of research avoids reliance on a pre-trained occupancy network~\cite{mohan2026forecastocc, li2025semi, zhang2026interocf, dang2026sparseworld, du2025sparseworld}.
ForecastOcc~\cite{mohan2026forecastocc} predicts the future development in the image feature space, while PreWorld~\cite{li2025semi} uses a self-supervised pre-training stage with 2D labels, followed by a fully supervised fine-tuning step.
InterOCF~\cite{zhang2026interocf} models spatio-temporal dynamics in 3D voxel and 2D segmentation space.
Building on a more efficient environment modeling, SparseWorld~\cite{dang2026sparseworld} leverages a sparse set of dynamic queries that directly interact with the image feature space. %
SparseWorld-TC~\cite{du2025sparseworld} adopts a similar idea but follows an action-conditioned design and predicts all future states in a single feedforward step.
Except for one variant of OccWorld, these approaches rely on human-labeled 3D or 2D ground truth labels and LiDAR depth, which our proposed approach does not require.
}
\section{Methodology}

Our approach for human label- and LiDAR-free semantic occupancy world model training, visualized in \Cref{fig:architecture}, consists of two stages.
In the first stage (\Cref{subsec:label_creation}), we generate 3D semantic occupancy pseudo-labels from unlabeled, posed multi-view images using vision foundation models by explicitly modeling long-term dynamics and completing vehicle shapes with a generative model.
Additionally, we leverage the ability to create voxel grids at multiple resolutions and with motion annotations.
In the second stage (\Cref{subsec:model_training}), we use these labels to train a sparse 3D semantic occupancy world model.
To improve training, we leverage higher-resolution pseudo-labels for fine-grained object classes and use the generated per-voxel motion annotations to focus training on dynamic scene parts while ensuring static consistency.

\subsection{Motion-Aware 3D Pseudo-Label Generation}\label{subsec:label_creation}
\subsubsection{3D Semantic Reconstruction}\label{subsubsec:recon}
First, we process multi-view images $\mathbf{I} \in \mathds{R}^{T \times M \times 3 \times H \times W}$ from $M$ views of a video with $T$ timestamps with a 4D feedforward reconstruction model $\phi$.
For our approach, we use MapAnything~\cite{keetha2026mapanything} and Pi3X~\citep{wang2026pi} since they accept camera poses $\mathbf{T}$ and intrinsics $\mathbf{K}$ as conditioning.
Concretely, $\phi$ outputs per-pixel depth maps $\mathbf{D} \in \mathds{R}^{T \times M \times H \times W}$ along with corresponding confidence values $\Omega \in [0,1]^{T \times M \times H \times W}$:
\begin{equation}
    \{\mathbf{D}, \Omega\} = \phi\!\left(\mathbf{I}, \mathbf{T}, \mathbf{K}\right),
\end{equation}
where $\mathbf{T} = \{\mathbf{T}_{t,m}\}_{t=1,m=1}^{T,M}$ with
$\mathbf{T}_{t,m} = \begin{bsmallmatrix} \mathbf{R}_{t,m} & \mathbf{t}_{t,m} \\ \mathbf{0}^\top & 1 \end{bsmallmatrix} \in \mathrm{SE}(3)$ being the recorded extrinsics for view $m$ at timestamp $t$ and $\mathbf{K} = \{\mathbf{K}_{m}\}_{m=1}^{M}$ with $\mathbf{K}_{m} \in \mathds{R}^{3\times 3}$ being the corresponding intrinsics of view $m$.
Since the estimated scale of the depth maps can be incorrect, especially for scenes with a static ego-vehicle, we compare it with the known camera height above the ground plane.
For each pixel $\mathbf{u} = (u, v)^\top$ in homogeneous coordinates $\tilde{\mathbf{u}} = (u, v, 1)^\top$, we lift its estimated depth into the global frame using the given camera parameters:
\begin{equation}
    \mathbf{P}^{\text{cam}}_{t,m}(\mathbf{u}) = \mathbf{D}_{t,m}(\mathbf{u})\, \mathbf{K}_{m}^{-1} \tilde{\mathbf{u}},
\end{equation}
\begin{equation}
    \mathbf{P}_{t,m}(\mathbf{u}) = \mathbf{R}_{t,m}\, \mathbf{P}^{\text{cam}}_{t,m}(\mathbf{u}) + \mathbf{t}_{t,m}.
\end{equation}

In parallel, we divide the semantic ground truth classes into potentially dynamic and non-dynamic ones.
For the classes that are always static, such as vegetation and road, we query an open-vocabulary semantic segmentation model $\psi_{\text{static}}$ (SAM3~\citep{carion2026sam}) with all images $\mathbf{I}$ and multiple prompts $\mathbf{Q}_{\text{static}}$ per class $c$ to retrieve segmentation masks $\mathbf{S}_{\text{static}, c, t}$ per class and timestamp $t$.
For each dynamic class $c'$, we use an open-vocabulary instance tracking model $\psi_{\text{dynamic}}$ (SAM3.1) to yield masks with semantics ($\mathbf{S}_{\text{dynamic}, c', t}$) and track IDs ($\mathbf{J}_t$), which we later use for motion segmentation.
We aggregate the masks and instance IDs from all classes via the highest confidence value per pixel.
In combination with the depth maps, we retrieve a semantic 3D point cloud.

\subsubsection{Static and Dynamic Disentanglement}\label{subsubsec:motion_seg}
{\parskip=2pt
\noindent\textbf{Long-Term Motion Segmentation.}
To determine which objects in potentially dynamic classes are actually moving, we design a long-term motion segmentation (LTMS) approach based on instance tracks and estimated depth maps at each timestamp $t$.
First, we compute the 3D centroid $\bar{\mathbf{P}}_{k,t}$ of each instance $k$ as the mean of its lifted pixels $\mathcal{U}_{k,t} = \left\{ \mathbf{u}: \mathbf{J}_{t}(\mathbf{u}) = k \right\}$ and use its ground plane coordinates $\mathbf{c}_{k,t}$, as we are only interested in the movement in BEV space:
{\small
\begin{equation}
\bar{\mathbf{P}}_{k,t} = \frac{1}{|\mathcal{U}_{k,t}|} \sum_{\mathbf{u} \in \mathcal{U}_{k,t}} \mathbf{P}_{k, t}(\mathbf{u}),
\qquad
\mathbf{c}_{k,t} = \Pi\, \bar{\mathbf{P}}_{k,t} \in \mathds{R}^2,
\end{equation}}
where $\Pi \in \mathds{R}^{2 \times 3}$ projects onto the ground plane.
Since these coordinates are already defined in the global coordinate frame, they are independent of the ego-motion, and we can associate them over time via the tracking IDs:
\begin{equation}
    \mathcal{C}_k = \left\{ \mathbf{c}_{k,t} \right\}_{t \in \mathcal{V}_k},
\qquad
\mathcal{V}_k \subseteq \{t_1, \dots, t_T\}.
\end{equation}
Since it can happen (especially in the side views) that the previous centroid of an object is no longer visible in the next frame, we remove the centroid prediction of this frame from $\mathcal{C}_k$.
To mitigate noise in the depth estimation, we then apply a forward and backward Kalman filter over time.
We exclude predictions that exceed a Mahalanobis distance threshold from the smoothed trajectory and iteratively re-run the smoothing.
We denote the smoothed BEV position of object $k$ at timestamp $t$ with $\tilde{\mathbf{c}}_{k,t}$.
To determine whether an object track is dynamic or static, we compute a distance-weighted average displacement $\mathrm{AD}_k$ per frame and the final displacement $\mathrm{FD}_k$:
\begin{equation}
\resizebox{0.95\columnwidth}{!}{$
\mathrm{AD}_k = \frac{\sum_{t=2}^{N_k} w_{k,t}
\lVert \tilde{\mathbf{c}}_{k,t} - \tilde{\mathbf{c}}_{k,t-1} \rVert_2}
{\sum_{t=2}^{N_k} w_{k,t}},
\quad
\mathrm{FD}_k = \lVert \tilde{\mathbf{c}}_{k,N_k} - \tilde{\mathbf{c}}_{k,1} \rVert_2
$},
\end{equation}
where $w_{k,t} = \left(\frac{d_{k,t-1} + d_{k,t}}{2}\right)^{-2}$ with $d_{k,t} = \lVert \bar{\mathbf{P}}_{k,t} - \mathbf{o}_t \rVert_2$ measures the distance between the centroid and the camera center $\mathbf{o}_t$ at frame $t$ and $N_k = \lvert \mathcal{V}_k \rvert$.
Since depth estimates closer to the camera are generally more reliable, this factor down-weights predictions farther from the ego-vehicle for $\mathrm{AD}_k$.
We define the thresholds relative to pre-defined speed thresholds $v_{\mathrm{AD}}$ and $v_{\mathrm{FD}}$:
\begin{equation}
\tau_{\mathrm{AD},k} = v_{\mathrm{AD}} \cdot \frac{1}{N_k - 1}\sum_{t=2}^{N_k} \left(s_{k,t} - s_{k,t-1}\right),
\end{equation}
\begin{equation}
    \tau_{\mathrm{FD},k} = v_{\mathrm{FD}} \cdot \left(s_{k,N_k} - s_{k,1}\right),
\end{equation}
with $s_{k,t}$ being the timestamp associated with $\mathbf{c}_{k,t}$,
and compare them to pre-defined thresholds:
\begin{equation}
    y_k =
\begin{cases}
\text{dynamic}, & \mathrm{AD}_k > \tau_{\mathrm{AD}, k} \;\wedge\; \mathrm{FD}_k > \tau_{\mathrm{FD}, k}, \\
\text{static}, & \text{otherwise}.
\end{cases}
\end{equation}
In practice, we also classify objects as dynamic if they have fewer than three valid predictions ($N_k < 3$) since the evidence for these objects is limited, and they could potentially pollute the static map.
For the same reason, we mark objects as dynamic if they exceed one of the two thresholds strongly. %
By employing this motion segmentation approach, we are able to accumulate not only the point clouds for the static classes per sequence but also the points of the non-moving objects, i.e., tracks with $y_k = \text{static}$, of potentially dynamic classes into the point cloud $\mathcal{P}_{\text{static}}$.
}

{\parskip=2pt
\noindent\textbf{Moving Object Completion.}\label{subsubsec:object-completion}
Since the dynamic point cloud $\mathcal{P}_{\text{dynamic}, t}$ per timestamp contains only a sparse point cloud of currently visible points for each moving object ($y_k = \text{dynamic}$), we use a generative model to complete the shape of objects from large vehicle classes (car, bus, trailer, and truck).
For our moving object completion (MOC), we employ the first stage of SAM 3D Objects~\citep{chen2026sam}, denoted by $\Theta$, to retrieve a set of occupied voxels for the shape-completed object $k$ at timestamp $t$ in a canonical occupancy grid $\mathcal{X}_{k,t} \subset \{0,\dots,63\}^3$. We provide an image crop $\mathbf{I}_{k,t}$ and the corresponding segmentation mask $\mathbf{J}_{t}(\mathbf{u}) = k$ as input and condition it on the observed point cloud $\mathbf{P}^{\text{cam}}_{k,t} = \left\{ \mathbf{P}^{\text{cam}}_{t,m}(\mathbf{u}) \right\}_{\mathbf{u} \in \mathcal{U}_{k,t}}$:
{\small
\begin{equation}
\left\{ \mathcal{X}_{k,t}, \mathbf{R}_{k,t}, \mathbf{t}_{k,t}, \sigma_{k,t} \right\} = \Theta\big(\mathbf{I}_{k,t}, \mathds{1}[\mathbf{J}_{t}(\mathbf{u}) = k], \mathbf{P}^{\text{cam}}_{k,t}\big).
\end{equation}}

Together with the voxel grid, the model also outputs rotation $\mathbf{R}_{k,t}$, translation $\mathbf{t}_{k,t}$, and scale $\sigma_{k,t}$ to place the object in the camera coordinate frame.
After using these parameters to transform the object and convert it to a point cloud, we insert these points into $\mathcal{P}_{\text{dynamic}, t}$.
We observe that the predicted scale $\sigma_{k, t}$ tends to be inaccurate, and we counteract this by reprojecting the object back onto the camera plane and computing an isotropic scale factor to match the corresponding 2D instance mask.
We perform shape completion only for large objects. For small objects such as pedestrians, the unobserved volume is much smaller, which limits the potential gain, while completing each of these numerous instances would add substantial computational cost.
}

\begin{table*}[htb]
\centering
\scriptsize
\setlength{\tabcolsep}{5pt}
\caption{4D occupancy forecasting performance on Occ3D-nuScenes~\cite{tian2023occ3d}, following the evaluation protocols of PreWorld~\cite{li2025semi} and SparseWorld-TC~\cite{du2025sparseworld}. Avg. denotes the averaged forecasting metrics over 1s, 2s, and 3s horizons. $\dagger$ means results reproduced by us with the released checkpoint. $\ddagger$ denotes that the method does not use 3D GT but depth supervision from LiDAR.}
\label{tab:occ_forecasting}
\begin{tabular}{c|l|cc|cccc|cccc|c}
\toprule
\multirow{2.5}{*}{Protocol} & \multirow{2.5}{*}{Method} & \multirow{2.5}{*}{2D GT} & \multirow{2.5}{*}{3D GT} & \multicolumn{4}{c}{Semantic mIoU (\%) $\uparrow$} & \multicolumn{4}{c}{Geometric IoU (\%) $\uparrow$} & \multirow{2.5}{*}{FPS $\uparrow$} \\
\cmidrule(lr){5-8} \cmidrule(lr){9-12}
& & & & 1s & 2s & 3s & Avg. & 1s & 2s & 3s & Avg. & \\
\midrule
& \multicolumn{12}{l}{\textit{Fully Supervised Methods}}\\
& OccWorld-D~\citep{zheng2024occworld}               & \xmark & \cmark & 11.55 & 8.10  & 6.22  & 8.62  & 18.90 & 16.26 & 14.43 & 16.53 & 2.8 \\
& PreWorld~\citep{li2025semi}                 & \cmark & \cmark & 12.27 & 9.24  & 7.15  & 9.55  & 23.62 & 21.62 & 19.63 & 21.62 & - \\
& SparseWorld~\citep{dang2026sparseworld}               & \xmark & \cmark & \underline{14.93} & \underline{13.15} & \underline{11.51} & \underline{13.20} & \underline{22.96} & \underline{22.10} & \underline{21.05} & \underline{22.03} & \underline{8.0} \\
& SparseWorld-TC-Small~\citep{du2025sparseworld}$\dagger$    & \xmark & \cmark & \textbf{17.77} & \textbf{16.23} & \textbf{14.79} & \textbf{16.26} & \textbf{24.60} & \textbf{23.87} & \textbf{22.91} & \textbf{23.79} & \textbf{9.35} \\
\cmidrule{2-13}
& \multicolumn{12}{l}{\textit{3D Label-Free Methods}}\\
& OccWorld-S~\citep{zheng2024occworld} & \xmark & \xmark & 0.28 & 0.26 & 0.24 & 0.26 & 5.05 & 5.01 & 4.95 & 5.00 & 2.8\\
& RenderWorld~\citep{yan2025renderworld} & \cmark & \xmark$^\ddagger$ & 2.83  & 2.55  & 2.37  & 2.58  & 14.61 & 13.61 & 12.98 & 13.73 & - \\
& PreWorld~\citep{li2025semi} & \cmark & \xmark$^\ddagger$ & 4.36 & 3.72 & 3.27 & 3.78 & 9.49 & 9.17 & 8.90 & 9.19 & -\\
\cmidrule{2-13}
\rowcolor{bannergray}
\cellcolor{white} & \approach (Ours) (+ SparseWorld~\cite{dang2026sparseworld}) & \xmark & \xmark & 6.66 & 5.94 & 5.30 & 5.97 & 15.00 & 14.64 & 14.26 & 14.63 & \underline{8.0}\\
\rowcolor{bannergray}
\cellcolor{white} & \approach (Ours) (+ SparseWorld-TC-Small~\cite{du2025sparseworld}) & \xmark & \xmark & \textbf{7.32} & \underline{6.79} & \underline{6.37} & \underline{6.83} & \underline{15.31} & \underline{15.07} & \underline{14.75} & \underline{15.04} & \textbf{9.35}\\
\rowcolor{bannergray}
\cellcolor{white}\multirow{-12}{*}{\rotatebox[origin=c]{90}{\textbf{PreWorld}}} & \approach (Ours) (+ SparseWorld-TC-Small*~\cite{du2025sparseworld}) & \xmark & \xmark & \underline{7.30} & \textbf{6.81} & \textbf{6.42} & \textbf{6.84} & \textbf{15.57} & \textbf{15.33} & \textbf{15.00} & \textbf{15.30} & 5.63\\
\midrule
& \multicolumn{12}{l}{\textit{Fully Supervised Methods}}\\
& PreWorld~\cite{li2025semi}$\dagger$ & \cmark & \cmark & 18.71 & 14.55 & 11.67 & 14.98 & 40.15 & 37.66 & 35.50 & 37.77 & -\\
& SparseWorld~\citep{dang2026sparseworld}$\dagger$               & \xmark & \cmark & 22.67 & 19.75 & 17.04 & 19.82 & 42.86 & 41.09 & 38.72 & 40.89 & \underline{8.0} \\
& SparseWorld-TC-Small~\citep{du2025sparseworld}    & \xmark & \cmark & \underline{27.95} & \underline{25.51} & \underline{23.35} & \underline{25.60} & \underline{50.69} & \underline{49.15} & \underline{47.23} & \underline{49.02} & \textbf{9.35} \\
& SparseWorld-TC-Small*~\cite{du2025sparseworld} & \xmark & \cmark & \textbf{30.09} & \textbf{27.61} & \textbf{25.43} & \textbf{27.71} & \textbf{52.66} & \textbf{51.20} & \textbf{49.31} & \textbf{51.05} & 5.63\\
\cmidrule{2-13}
& \multicolumn{12}{l}{\textit{3D Label-Free Methods}}\\
\rowcolor{bannergray}
\cellcolor{white} & \approach (Ours) (+ SparseWorld~\cite{dang2026sparseworld}) & \xmark & \xmark & 13.49 & 12.09 & 10.73 & 12.10 & 39.10 & 37.78 & 36.03 & 37.64 & \underline{8.0}\\
\rowcolor{bannergray}
\cellcolor{white} & \approach (Ours) (+ SparseWorld-TC-Small~\cite{du2025sparseworld}) & \xmark & \xmark & \underline{15.72} & \underline{14.73} & \underline{13.98} & \underline{14.81} & \underline{43.17} & \underline{42.46} & \underline{41.48} & \underline{42.37} & \textbf{9.35}\\
\rowcolor{bannergray}
\cellcolor{white}\multirow{-9}{*}{\rotatebox[origin=c]{90}{\textbf{SparseWorld-TC}}} & \approach (Ours) (+ SparseWorld-TC-Small*~\cite{du2025sparseworld}) & \xmark & \xmark & \textbf{15.97} & \textbf{15.04} & \textbf{14.29} & \textbf{15.10} & \textbf{44.53} & \textbf{43.83} & \textbf{42.78} & \textbf{43.71} & 5.63\\
\bottomrule
\end{tabular}
\vspace{-0.3cm}
\end{table*} %

\subsubsection{Motion-Aware Resolution-Adaptive Voxelization}
\label{subsec:voxelization}
Given the combined semantic point cloud $\mathcal{P}_t$ ($\mathcal{P}_{\text{static}} \cup \mathcal{P}_{\text{dynamic}, t}$) at timestamp $t$ and a pre-defined voxel grid, we voxelize the points following ShelfOcc~\cite{boeder2025shelfocc} and FreeOcc~\cite{caunes2026freeocc}.
We provide more details about the voxelization in the supplementary material \Cref{sec:supp_voxelization}.
In contrast to these works, and to better learn fine details of objects while still handling large stuff classes during downstream training efficiently, we generate our pseudo-labels at two different resolutions: a coarse $\mathbf{V}_{t, \text{coarse}}$, following the Occ3D~\cite{tian2023occ3d} voxel size, and a fine-grained $\mathbf{V}_{t, \text{fine}}$, with doubled resolution.
Complementary to this, we store a dynamic mask $\mathbf{M}_{t, \text{dynamic, coarse/fine}}$ of the same shape as $\mathbf{V}_{t, \text{coarse/fine}}$ that evaluates to $1$ whenever at least one point from $\mathcal{P}_{\text{dynamic}, t}$ falls into a voxel and $0$ otherwise, and which we use to guide world model training.

\subsection{World Model Training}\label{subsec:model_training}
Our underlying world model~\cite{du2025sparseworld} $\mathcal{F}$ takes $M$ multi-view images $\mathbf{I} \in \mathds{R}^{T' \times M \times 3 \times H \times W}$ from the past $T'$ timestamps, the conditioning trajectory $\tau \in \mathds{R}^{T_F \times d'}$ for the future $T_F$ timestamps, and initial query states $\mathcal{M}_{1:T_F}$ as input to forecast future occupancy $\mathcal{O}_{1:T_F}$:
\begin{equation}
    \mathcal{O}_{1:T_F} = \mathcal{F}(\mathcal{M}_{1:T_F}, \mathbf{I}, \tau).
\end{equation}
Each waypoint of $\tau$ comprises the relative pose between the future and the current ego pose, and its timestamp.
Internally, the world model first extracts image features through an image backbone and a neck.
The sparse transformer decoder operates on a set of $N_Q$ queries, each consisting of a $d$-dimensional feature vector and a set of 3D points that increases over decoder layers.
To integrate the action conditioning, it adds a timestamp encoding to all queries and modulates them with the relative pose between the future waypoint and the current pose.
In each decoder layer, every query predicts 3D sampling points, which are transformed into the current ego frame and projected into all past image feature maps to perform deformable attention and aggregate features via adaptive mixing.
Next, all queries interact via temporal self-attention across frames, per-frame self-attention, and a feed-forward network, before two MLPs predict per-point semantic logits and 3D offsets.

{\parskip=2pt
\noindent\textbf{Hybrid Resolution Supervision.}
To derive the training supervision in the form of the point cloud $\mathds{P}_t$ for every timestamp $\mathcal{T} = \{t_\text{curr}, \dots, t_\text{curr} + T_F\}$ up to the forecasting horizon $T_F$, we utilize the two different resolutions of our generated pseudo-labels to create a hybrid resolution supervision (HRS).
For large stuff classes, we extract the center point of all voxels belonging to these classes and their semantic label from the coarse grid $\mathbf{V}_{t, \text{coarse}}$, while we use $\mathbf{V}_{t, \text{fine}}$ for the object classes.
This reduces information loss during voxelization for small classes and allows us to represent object classes, such as pedestrians, with more fine-grained details and surfaces more accurately.
Employing our proposed hybrid resolution enables this representation with minimal overhead as we still represent most of the scene at a coarse resolution. %
}

{\parskip=2pt
\noindent\textbf{Motion-Guided Losses.}
During loss calculation, we use the second advantageous property of our pseudo-labels: the inclusion of the dynamic mask $\mathbf{M}_{t, \text{dynamic, coarse/fine}}$ per voxel, through motion-guided losses (MGL).
For simplicity, we denote the dynamic label of the pseudo-label supervision point $\mathbf{p}$  with $\mathbf{M}_{t,\text{dynamic}}(\mathbf{p})$, independent of the voxel grid resolution.
Since moving objects are the most important but also the most difficult scene parts to learn for a world model due to their infrequent occurrence, we use the dynamic mask to up-weight points from dynamic voxels with weight factor $\gamma$.
For all timestamps up to the forecasting horizon, we compute an adaptation of the Chamfer distance as the loss
function for the predicted point clouds $\{\mathds{\hat{P}}_0, \dots, \mathds{\hat{P}}_{T_F}\}$:}
\begin{equation}
\mathcal{L}_{\text{CD}} = \sum_{t \in \mathcal{T}} \Big(
  D_{w}(\mathds{P}_t, \hat{\mathds{P}}_t)
  + D_{1}(\hat{\mathds{P}}_t, \mathds{P}_t) \Big),
\end{equation}
where
\begin{equation}
D_{w}(\mathds{X}, \mathds{Y}) = \frac{1}{\lvert\mathds{X}\rvert}
  \sum_{\mathbf{x} \in \mathds{X}} w(\mathbf{x})
  \min_{\mathbf{y} \in \mathds{Y}} \lVert \mathbf{x} - \mathbf{y} \rVert_1
\end{equation}
is the weighted directional Chamfer distance with
$w(\mathbf{p}) = 1 + (\gamma - 1)\,\mathbf{M}_{t,\text{dynamic}}(\mathbf{p})$,
and $D_{1}$ denotes its unweighted version.
We supervise the semantic class predictions with the focal loss $\mathcal{L}_{\text{focal}}$, in which we also up-weight points from dynamic voxels with the same weight as for the geometry.%

Further, we add a static consistency
loss. Let $\mathds{A}$ be the pseudo-label points of the current frame at
non-dynamic voxels with $\mathbf{M}_{0,\text{dynamic}}(\mathbf{p})=0$.
For all future timestamps $t \in \mathcal{T}_F = \mathcal{T}\setminus\{t_\text{curr}\}$, we transform the predicted point cloud $\mathds{\hat{P}}_t$ into the current ego frame (denoted by $\tilde{\mathds{P}}_t$).
For the subset of static pseudo-label points $\mathds{A}_t \subseteq \mathds{A}$ still within the observed grid at time $t$, we apply a
one-directional Chamfer distance from the static pseudo-labels to the predictions:
\begin{equation}
\mathcal{L}_{\text{static}} = \frac{1}{|\mathcal{T}_F|} \sum_{t \in \mathcal{T}_F} D_1(\mathds{A}_t, \tilde{\mathds{P}}_t),
\end{equation}
which yields the overall loss with the static loss weight $\eta$:
\begin{equation}
\mathcal{L} = \mathcal{L}_{\text{CD}} + \mathcal{L}_{\text{focal}} + \eta\,\mathcal{L}_{\text{static}}.
\end{equation}

\section{Experiments}
\subsection{Experimental Setup}
{\noindent\textbf{Datasets and Metrics.}
We conduct our experiments on the Occ3D-nuScenes and Occ3D-Waymo benchmark~\cite{tian2023occ3d}, building on the nuScenes~\cite{caesar2020nuscenes, fong2022panoptic} and Waymo Open Dataset~\cite{sun2020scalability}.
Following established evaluation protocols, we report the geometric Intersection over Union (IoU), semantic mean Intersection over Union (mIoU) (excluding \textit{others} and \textit{other\_flat}), and RayIoU~\cite{liu2024fully} for 3D semantic occupancy prediction in \% on Occ3D-nuScenes.
For 4D semantic occupancy forecasting, we report mIoU and IoU in \% at forecasted horizons of 1s, 2s, and 3s, using the average as the main metric.
Here, we do not exclude any classes to ensure direct comparability with fully supervised methods and published baselines.
We observe that previous works differ in their evaluation protocol.
PreWorld~\cite{li2025semi} and SparseWorld~\cite{dang2026sparseworld} evaluate all voxels across all frames with a past horizon of five and a future horizon of six frames (4219 frames total), whereas the released SparseWorld-TC~\cite{du2025sparseworld} code evaluates only camera-visible voxels, following the occupancy prediction standard. In addition, its evaluated set shrinks along the predicted horizon from 6019 at t+0 to 5119 at t+3 s.
We therefore report both for nuScenes, and re-evaluate the released model checkpoints of PreWorld, SparseWorld, and SparseWorld-TC under both protocols.
For Waymo and all ablations, we use the SparseWorld-TC protocol.
}

\begin{table}[t]
\centering
\scriptsize
\setlength{\tabcolsep}{2pt}
\caption{4D occupancy forecasting performance on Occ3D-Waymo~\cite{tian2023occ3d}, following the evaluation protocol of SparseWorld-TC~\cite{du2025sparseworld}. Avg. denotes the averaged forecasting metrics over 1s, 2s, and 3s horizons. SW-TC represents SparseWorld-TC-Small.}
\label{tab:occ_forecasting_waymo}
\begin{tabular}{l|cc|cccc|cccc}
\toprule
\multirow{2.5}{*}{Method} & \multirow{2.5}{*}{\shortstack{2D\\GT}} & \multirow{2.5}{*}{\shortstack{3D\\GT}} & \multicolumn{4}{c}{Semantic mIoU (\%) $\uparrow$} & \multicolumn{4}{c}{Geometric IoU (\%) $\uparrow$} \\
\cmidrule(lr){4-7} \cmidrule(lr){8-11}
& & & 1s & 2s & 3s & Avg. & 1s & 2s & 3s & Avg.\\
\midrule
\multicolumn{11}{l}{\textit{Fully Supervised Methods}}\\
PreWorld~\citep{li2025semi}                 & \cmark & \cmark & 11.09 & 8.97 & 7.80 & 9.29 & 40.19 & 35.25 & 31.99 & 35.81\\
SW-TC~\citep{du2025sparseworld}    & \xmark & \cmark & \textbf{17.50} & \textbf{15.54} & \textbf{14.12} & \textbf{15.72} & \textbf{47.34} & \textbf{44.98} & \textbf{42.77} & \textbf{45.03}\\
\midrule
\multicolumn{11}{l}{\textit{3D Label-Free Methods}}\\
PreWorld~\citep{li2025semi} & \cmark & \xmark & 7.37 & 6.74 & 6.20 & 6.77 & 29.15 & 29.62 & 29.33 & 29.37\\
\midrule
\rowcolor{bannergray}
Ours (+SW-TC) & \xmark & \xmark & \textbf{8.50} & \textbf{8.09} & \textbf{7.76} & \textbf{8.12} & \textbf{30.90} & \textbf{30.50} & \textbf{30.13} & \textbf{30.51}\\ %
\bottomrule
\end{tabular}
\vspace{-0.3cm}
\end{table} %
\begin{table}
\centering
\scriptsize
\caption{3D semantic occupancy estimation performance on Occ3D-nuScenes. We compare our method against previous works. $\dagger$~denotes that the method uses STCOcc~\cite{liao2025stcocc} as occupancy model.}
\label{tab:rayiou_results}
\setlength{\tabcolsep}{1.65pt}
\begin{tabular}{l|cccc|c}
\toprule
Method & RayIoU$\uparrow$ & RayIoU@1$\uparrow$ & RayIoU@2$\uparrow$ & RayIoU@4$\uparrow$ & mIoU$\uparrow$\\
\midrule
SelfOcc~\cite{huang2024selfocc} & 10.5 & 8.9 & 10.5 & 12.0 & 10.54\\
GaussianOcc~\citep{gan2025gaussianocc} & 11.85 & 8.69  & 11.90 & 14.95 & 11.26\\
TT-Occ-Camera~\cite{zhang2026test} & 13.4 & 10.0 & 12.9 & 15.7 & 16.70\\
GaussianFlowOcc~\citep{boeder2025gaussianflowocc} & 16.47 & 11.81 & 16.58 & 20.98 & 17.08\\
ShelfOcc$^\dagger$~\citep{boeder2025shelfocc} & 19.97 & 14.38 & 20.07 & 25.47  & 22.87\\
QueryOcc~\cite{lilja2026queryocc} & 20.8 & - & - & - & 21.3\\
LetOccVote-L~\cite{zhang2026letoccvote} & 20.91 & \underline{15.80} & \underline{21.30} & 25.62 & \underline{23.11}\\
FreeOcc$^\dagger$~\citep{caunes2026freeocc} & \underline{21.1} & 13.6 & \underline{21.3} & \textbf{28.5} & 22.8\\
\midrule
\rowcolor{bannergray}
\approach (Ours)$^\dagger$ & \textbf{23.75} & \textbf{18.52} & \textbf{24.38} & \underline{28.36} & \textbf{24.86}\\
\bottomrule
\end{tabular}
 \vspace{-0.3cm}
\end{table}

{\parskip=2pt
\noindent\textbf{Implementation Details.}
Following ShelfOcc~\cite{boeder2025shelfocc} and FreeOcc~\cite{caunes2026freeocc}, we evaluate the quality of our pseudo-labels by training STCOcc~\cite{liao2025stcocc} with a ResNet-50~\cite{he2016deep} backbone and an image resolution of $256 \times 704$ and compare against other weakly- and self-supervised approaches.
For 4D occupancy forecasting, we train SparseWorld~\cite{dang2026sparseworld} for 54 and SparseWorld-TC-Small~\cite{du2025sparseworld} for 70 epochs with a ResNet-50 backbone, in accordance with the respective papers, as world models.
For nuScenes, we additionally report results for SparseWorld-TC-Small* with a ConvNeXt-Base~\cite{liu2022convnet} backbone, initialized with DINOv3~\cite{simeoni2026dinov} weights.
For all ablation experiments, we use SparseWorld-TC-Small and train for 20 epochs.
We use Pi3X~\cite{wang2026pi} during label creation if not otherwise noted.
More details, including the prompts used for SAM3 and pseudo-label generation time, are detailed in the supplementary material \Cref{sec:supp_implementation}.
}

\subsection{Quantitative Results}
\noindent\textbf{4D Occupancy Forecasting.}
We evaluate \approach on Occ3D-nuScenes in \Cref{tab:occ_forecasting}, following the PreWorld and SparseWorld-TC evaluation protocols.
The results of the fully supervised methods, which utilize 3D human-labeled ground truth during training, serve as reference values.
Across different underlying sparse world models, including SparseWorld and two variations of SparseWorld-TC-Small, we significantly surpass previous 3D-label-free methods by at least 57.9\% in average forecasted mIoU (5.97 vs. 3.78) and up to 81.0\% (6.84 vs. 3.78) for the PreWorld evaluation protocol, despite not using 2D ground truth labels or LiDAR depth supervision as the strongest comparison methods do.
When only evaluated in the visible area (SparseWorld-TC protocol), \approach combined with SparseWorld-TC-Small* performs on par with the fully supervised method PreWorld in mIoU (15.10 vs. 14.98) and IoU (43.71 vs. 37.77), demonstrating the potential of label-free training.
However, compared to the fully supervised versions of SparseWorld and SparseWorld-TC trained with human-annotated and LiDAR-derived 3D voxel grids, there still exists a gap in mIoU (15.10 vs. 27.71 for SparseWorld-TC-Small*) and IoU (43.71 vs. 51.05 for SparseWorld-TC-Small*).
We show quantitative results for Occ3D-Waymo in \Cref{tab:occ_forecasting_waymo}.
Here, we also observe that \approach surpasses the 2D ground truth supervised version of PreWorld and lies closer in mIoU points to its 3D-supervised variant.
We provide additional results in \Cref{sec:supp_quantitative}. %

\begin{table}[t]
\scriptsize
\centering
\caption{Quality of the 3D semantic occupancy pseudo-labels.\label{tab:pseudo_label}}
\begin{tabular}{lc|ccc} 
    \toprule
    Method & 3D Model & mIoU $\uparrow$ & IoU $\uparrow$ & RayIoU $\uparrow$\\
    \midrule
    ShelfOcc~\citep{boeder2025shelfocc} & MapAnything & 9.6 & 17.2 & -\\
    FreeOcc~\citep{caunes2026freeocc} & MapAnything & 20.4 & - & 16.8\\
    \midrule
    \rowcolor{bannergray}
    \approach (Ours) & MapAnything & \underline{22.4} & \underline{53.0} & \underline{21.5}\\
    \rowcolor{bannergray}
    \approach (Ours) & Pi3X & \textbf{22.9} & \textbf{54.1} & \textbf{22.0}\\
    \bottomrule
\end{tabular}
\end{table}

\begin{table}[t]
\scriptsize
\centering
\caption{Ablation study for the individual components. All values are averaged over the 1\,s, 2\,s, and 3\,s forecasting horizons. mIoU$_\text{dyn}$ includes all potentially moving classes. mIoU$_\text{mov}$ and mIoU$_\text{sta}$ evaluate the voxels from moving and static objects of these classes, determined via the motion attributes of the GT bounding boxes.\label{tab:abl_main}}
\setlength{\tabcolsep}{3.5pt}
\begin{tabular}{cccc|cccc}
\toprule
LTMS & \shortstack{MOC} & \shortstack{HRS} & \shortstack{MGL} & \shortstack{mIoU $\uparrow$} & \shortstack{mIoU$_\text{dyn}$ $\uparrow$} & \shortstack{mIoU$_\text{mov}$ $\uparrow$} & \shortstack{mIoU$_\text{sta}$ $\uparrow$}\\
    \midrule
    & & & & 11.61 & 2.01 & 0.55 & 3.20\\
    \cmark & & & & 12.33 & 3.50 & 0.62 & 6.38\\
    \cmark & \cmark & & & 12.54 & 4.02 & 1.05 & 7.23\\
    \cmark & \cmark & \cmark & & 13.74 & 5.78 & 1.98 & 10.47\\
    \rowcolor{bannergray}
    \cmark & \cmark & \cmark & \cmark & \textbf{14.04} & \textbf{6.89} & \textbf{4.00} & \textbf{12.23}\\
    \bottomrule
\end{tabular}
 \vspace{-0.35cm}
\end{table}

\begin{figure*}[t]
    \centering
    {\includegraphics[width=0.8\linewidth]{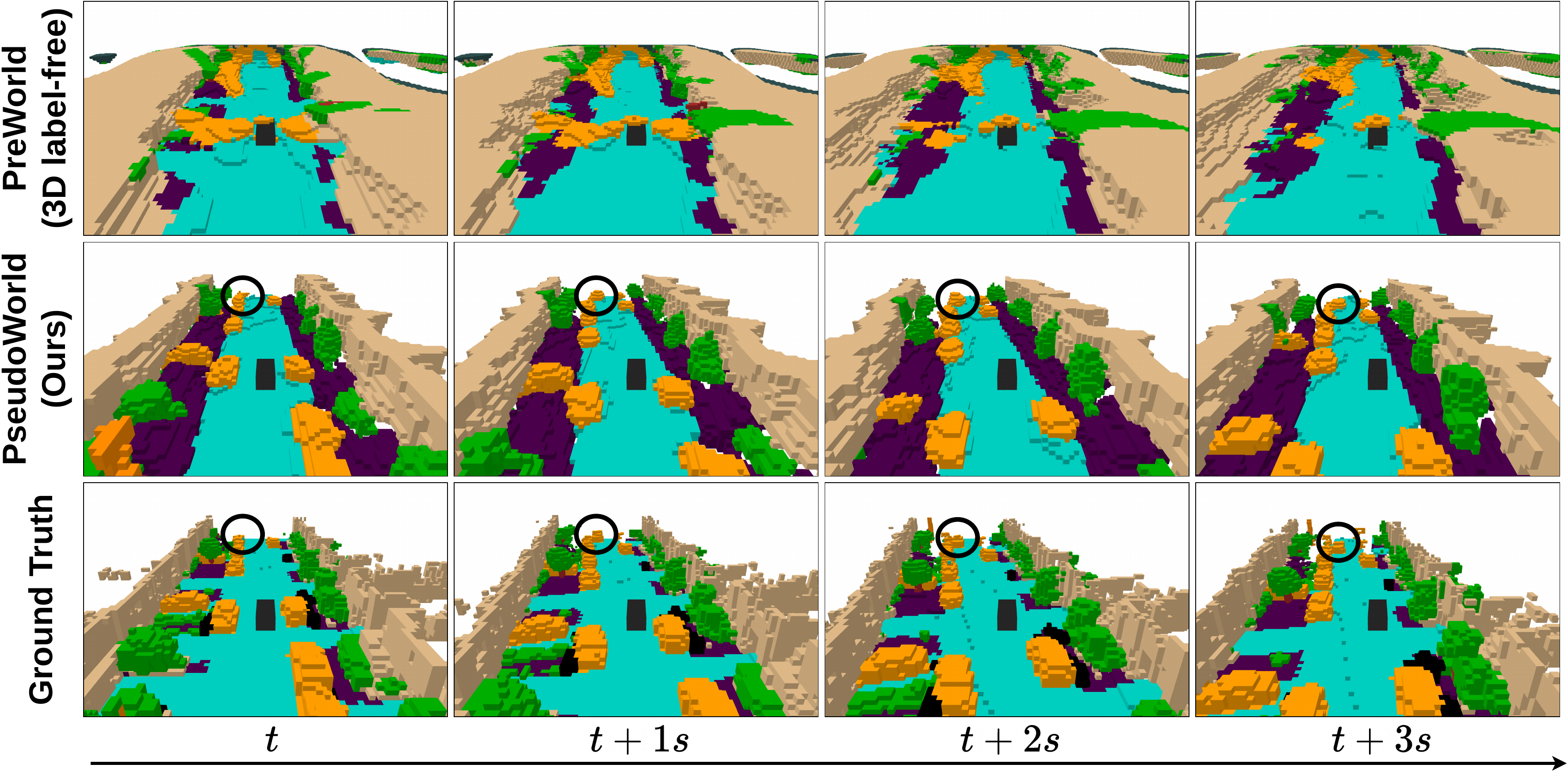}}
    \caption{Qualitative results on the Occ3D-Waymo dataset. Compared with the 3D label-free version of PreWorld~\cite{li2025semi}, \approach exhibits less depth bleeding, which PreWorld's 2D rendering supervision causes, and correctly predicts the parked cars as static while the ego-vehicle passes by. It also recognizes and forecasts the motion of the oncoming car.}
    \label{fig:qualitative}
    \vspace{-0.3cm}
\end{figure*}

{\parskip=2pt
\noindent\textbf{3D Occupancy Prediction and Label Quality.}
As shown in \Cref{tab:rayiou_results}, \approach achieves state-of-the-art performance in terms of RayIoU (23.75 vs. previous best 21.1 from FreeOcc~\cite{caunes2026freeocc}) and mIoU (24.86 vs. previous best 23.11 from LetOccVote-L~\cite{zhang2026letoccvote}), and performs especially well for tolerances within 1m (RayIoU@1), demonstrating high geometric and semantic accuracy only from camera-derived, automatically generated labels.
When directly evaluating our pseudo-labeling framework on the Occ3D-nuScenes validation set in \Cref{tab:pseudo_label}, we also achieve an improvement in mIoU, IoU, and especially RayIoU (21.5 and 22.0 vs. 16.8 for FreeOcc) compared to ShelfOcc and FreeOcc with two different 3D foundation models.
}

\subsection{Ablation Study}
\noindent\textbf{Model Components.}
We study the influence of our proposed core components in \Cref{tab:abl_main}.
The first row represents training the underlying world model with pseudo-labels generated by determining dynamic objects based on semantics, similarly to ShelfOcc and FreeOcc, which do not yet provide publicly available code or labels.
Adding our proposed LTMS improves the mIoU by 0.72 and mIoU$_\text{dyn}$ for potentially dynamic classes by 1.49.
This is primarily attributed to the accumulation of stationary objects, which greatly increases their score from 3.20 to 6.38 (mIoU$_\text{sta}$).
Using MOC improves all metrics, including the mIoU of actually moving objects (mIoU$_\text{mov}$), but the effect is limited, as it applies to only four classes.
Deriving the HRS further boosts mIoU by 1.20 and for dynamic classes by 1.76, benefiting both moving and static objects.
Finally, our proposed MGL of the world model increases mIoU and especially enhances the performance of moving objects (from 1.98 to 4.00).
We provide more ablations in \Cref{sec:supp_quantitative}. %

\begin{table}[t]
    \scriptsize
    \centering

    \begin{minipage}{0.54\linewidth}
        \centering
        \setlength{\tabcolsep}{1pt}
        \caption{Ablation study for the motion segmentation algorithm.\label{tab:abl_ttocc}}
        \begin{tabular}{l|ccc} 
            \toprule
            Algorithm & mIoU $\uparrow$ & mIoU$_\text{dyn}$ $\uparrow$ & RayIoU$\uparrow$\\
            \midrule
            TT-Occ~\cite{zhang2026test} & 22.48 & 12.27 & 21.12\\
            \rowcolor{bannergray}
            Ours & \textbf{22.94} & \textbf{13.37} & \textbf{22.03}\\
            \bottomrule
        \end{tabular}
    \end{minipage}\hfill
    \begin{minipage}{0.44\linewidth}
        \centering
        \setlength{\tabcolsep}{3pt}
        \caption{Ablation study for AD \& FD thresholds.
        \label{tab:abl_motion_seg}}
        \begin{tabular}{cc|cc} 
            \toprule
            AD & FD & mIoU $\uparrow$ & RayIoU $\uparrow$\\
            \midrule
            \xmark & \cmark & 22.35 & 21.23\\
            \cmark & \xmark & 22.34 & 21.18\\
            \rowcolor{bannergray}
            \cmark & \cmark & \textbf{22.94} & \textbf{22.03}\\
            \bottomrule
        \end{tabular}
    \end{minipage}
\end{table}

\begin{table}[t]
    \scriptsize
    \centering

    \begin{minipage}{0.48\linewidth}
        \centering
        \caption{Ablation study for the motion loss factor $\gamma$.\label{tab:abl_motion_weight}}
        \setlength{\tabcolsep}{2.5pt}
        \begin{tabular}{c|cc} 
            \toprule
            $\gamma$ & Avg. mIoU $\uparrow$ & Avg. mIoU$_{\text{mov}}$ $\uparrow$\\
            \midrule
            1.0 & 13.75 & 2.33\\
            2.0 & 13.75 & 2.61\\
            \rowcolor{bannergray}
            5.0 & \textbf{14.04} & 4.00\\
            10.0 & 13.55 & \textbf{4.52}\\
            \bottomrule
        \end{tabular}
        \vspace{-0.2cm}
    \end{minipage}\hfill
    \begin{minipage}{0.48\linewidth}
        \centering
        \caption{Ablation study for the static consistency loss factor $\eta$.\label{tab:abl_consistency_weight}}
        \setlength{\tabcolsep}{2.5pt}
        \begin{tabular}{c|cc} 
            \toprule
            $\eta$ & Avg. mIoU $\uparrow$ & Avg. mIoU$_{\text{stuff}}$ $\uparrow$\\
            \midrule
            0 & 13.92 & 18.95\\
            1.0 & 14.03 & 19.06\\
            \rowcolor{bannergray}
            2.0 & \textbf{14.04} & \textbf{19.09}\\
            5.0 & 13.77 & 18.87\\
            \bottomrule
        \end{tabular}
        \vspace{-0.2cm}
    \end{minipage}
    
\end{table}

{\parskip=2pt
\noindent\textbf{Motion Segmentation.}
In \Cref{tab:abl_ttocc}, we compare the quality of our 3D pseudo-labels when using our motion segmentation strategy against the motion segmentation via optical flow from TT-Occ-Camera~\cite{zhang2026test}.
Since the latter only considers motion between neighboring frames and struggles more with distant objects, \approach improves mIoU by more than 1 point for dynamic object classes.
In \Cref{tab:abl_motion_seg}, we ablate the influence of the proposed AD and FD for track classification.
We observe that combining both yields the best label quality.
}

{\parskip=2pt
\noindent\textbf{Loss Factors.}
We ablate different values for the motion loss weight factor $\gamma$ during world model training in \Cref{tab:abl_motion_weight}.
While disabling the up-weighting ($\gamma = 1.0$) scores the lowest for mIoU$_\text{mov}$, increasing its weight improves it but negatively impacts the overall mIoU for values larger than $5$, as static parts get less weight.
In \Cref{tab:abl_consistency_weight}, we evaluate the performance of different static consistency loss factors $\eta$ and observe that values of $1$ and $2$ yield an increase in the overall mIoU and mIoU$_\text{stuff}$ for non-dynamic classes.
However, values greater than $2$ deteriorate mIoU due to worse results for moving objects.
}

\subsection{Qualitative Results}

In \Cref{fig:qualitative}, we visually compare \approach with the 3D-label-free variant of PreWorld~\cite{li2025semi} on Occ3D-Waymo.
In contrast to PreWorld, our method exhibits significantly less depth bleeding as it does not rely on 2D rendering supervision.
Moreover, our model demonstrates that it has learned to discriminate between static and moving objects. It correctly detects parked cars and recognizes and forecasts the motion of the distant, oncoming car.
We present more qualitative examples in the supplementary material \Cref{sec:supp_qualitative}.
\section{Conclusion}
In this work, we presented \approach, a novel framework for label-free 3D semantic occupancy world model training solely based on posed multi-view images.
We proposed a 3D semantic occupancy pseudo-labeling approach that explicitly determines moving objects via their long-term displacements and completes the shape of moving vehicles.
By designing a hybrid-resolution-based training mechanism that uses the generated labels and motion annotations to favor moving objects and enforce static consistency, we demonstrated significantly improved performance compared to previous approaches.

{\parskip=2pt
\noindent\textbf{Limitations.}
A limitation of our approach is the reliance on known camera poses and intrinsics. Future work could further investigate directly using metrically estimated poses from feedforward foundation models.
Moreover, while our approach showed promising performance without any human-labeled samples, we did not assess its scaling performance for potential deployment on raw fleet data.
}

\section*{LLM Usage Statement}
OpenAI ChatGPT and Anthropic Claude were used to assist with writing refinement and code development. The authors reviewed all AI-assisted content and take full responsibility for the submitted work.
\section*{Acknowledgement}
This research has been conducted as part of the HIDDEN project, which is funded by the European Union, under grant agreement No 101202228. Views and opinions expressed are however those of the author(s) only and do not necessarily reflect those of the European Union or the European Climate, Infrastructure and Environment Executive Agency (CINEA). Neither the European Union nor the granting authority can be held responsible for them.
Additionally, this work was funded by an academic grant from NVIDIA.
{
    \small
    \bibliographystyle{ieeenat_fullname}
    \bibliography{main}
}

\clearpage
\maketitlesupplementary

\appendix
\counterwithin{figure}{section}
\counterwithin{table}{section}

\section{Additional Implementation Details}\label{sec:supp_implementation}

\subsection{Datasets}
{\noindent\textbf{Occ3D-nuScenes}~\cite{tian2023occ3d} provides dense 3D semantic occupancy labels for the 700 training and 150 validation scenes of nuScenes, recorded with six surround-view cameras.
It contains labeled keyframes at 2 Hz (28{,}130 training and 6{,}019 validation frames) and covers a region of $[-40, 40]\,\text{m} \times [-40, 40]\,\text{m} \times [-1, 5.4]\,\text{m}$ around the ego vehicle at a voxel size of 0.4~m, resulting in a $200 \times 200 \times 16$ grid resolution.
Each voxel is assigned one of 17 semantic classes (others, barrier, bicycle, bus, car, construction vehicle, motorcycle, pedestrian, traffic cone, trailer, truck, driveable surface, other flat, sidewalk, terrain, manmade, vegetation) or free.
For mIoU$_\text{dyn}$, we compute the mean over the classes bicycle, bus, car, construction vehicle, motorcycle, pedestrian, trailer, and truck.
For mIoU$_\text{mov}$ and mIoU$_\text{sta}$, we split the voxels of these eight classes into moving and static objects using the nuScenes box attributes.
Vehicles and pedestrians are counted as moving if they have the attribute \textit{moving} at $t$ or $t{+}k$ and static otherwise. For bicycles and motorcycles, we consider them moving if they have the attribute \textit{rider} and their box center moves more than 0.4 m between $t$ and $t{+}k$.
mIoU$_\text{stuff}$ computes the mean IoU over the remaining classes (others, barrier, traffic cone, driveable surface, other flat, sidewalk, terrain, manmade, vegetation).
}

{\parskip=2pt
\noindent\textbf{Occ3D-Waymo}~\cite{tian2023occ3d} provides annotations with the same grid resolution and size for the 798 training and 202 validation sequences of the Waymo Open Dataset with five cameras.
For Waymo, annotations are available at 10 Hz (158{,}081 training and 39{,}987 validation frames), using the same spatial range and voxel size.
It distinguishes 15 semantic classes (general object, vehicle, pedestrian, sign, bicyclist, traffic light, pole, construction cone, bicycle, motorcycle, building, vegetation, tree trunk, road, walkable) and free.
}

\definecolor{rowshade}{gray}{0.92}

\begin{table}[t]
\centering
\scriptsize
\caption{SAM3 and SAM3.1 text prompts per class for Occ3D-nuScenes and Occ3D-Waymo. Rows are aligned by
semantics between the two datasets. $^\dagger$ denotes dynamic classes used for LTMS with SAM3.1, while all other classes are segmented with SAM3. We exclude ambiguous classes (\textit{others}, \textit{other\_flat}, and \textit{general object (GO)}). Sky is used to exclude points from the sky for point cloud generation.}
\label{tab:sam3_prompts}
\setlength{\tabcolsep}{3pt}
\renewcommand{\tabularxcolumn}[1]{m{#1}}
\begin{tabularx}{\columnwidth}{l>{\raggedright\arraybackslash}Xl>{\raggedright\arraybackslash}X}
\toprule
\multicolumn{2}{c}{nuScenes} & \multicolumn{2}{c}{Waymo} \\
\cmidrule(lr){1-2}\cmidrule(lr){3-4}
Class & Prompts & Class & Prompts \\
\midrule
\rowcolor{rowshade} car$^\dagger$        & car, sedan, van     &                                     & car, sedan, van \\
\rowcolor{rowshade} bus$^\dagger$        & bus                 &                                     & bus \\
\rowcolor{rowshade} truck$^\dagger$      & lorry, truck        &                                     & truck, lorry \\
\rowcolor{rowshade} trailer$^\dagger$    & trailer             & \multirow{-4.5}{*}{vehicle$^\dagger$} & trailer \\
cons. vehicle                            & excavator, crane    & --                                  & -- \\
\rowcolor{rowshade} pedestrian$^\dagger$ & person, pedestrian  & pedestrian$^\dagger$                & person, pedestrian \\
bicycle$^\dagger$                        & bicycle             & bicycle$^\dagger$                   & bicycle \\
\rowcolor{rowshade} --                   & --                  & bicyclist$^\dagger$                 & cyclist, bicyclist \\
motorcycle$^\dagger$                     & motorcycle, scooter & motorcycle$^\dagger$                & motorcycle, scooter \\
\rowcolor{rowshade} traffic cone         & traffic-cone        & cons. cone                          & traffic-cone \\
barrier                                  & barricade, barrier  & --                                  & -- \\
\midrule
\rowcolor{rowshade} driv. surface        & highway, road, roadmarking & road                         & highway, road, roadmarking \\
sidewalk                                 & sidewalk, walkway   &                                     & sidewalk, walkway, stairs \\
terrain                                  & turf, grass, sand   & \multirow{-2}{*}{walkable}          & turf, grass \\
\rowcolor{rowshade}                      & building, wall, fence, bridge, billboard & building       & building, wall, bridge \\
\rowcolor{rowshade}                      & sign                & sign                                & sign, traffic-sign \\
\rowcolor{rowshade}                      & light               & traffic light                       & traffic-light \\
\rowcolor{rowshade} \multirow{-4}{*}{manmade} & pole           & pole                                & pole \\
                                         &                     & vegetation                          & bush, plants, tree, flower \\
\multirow{-2}{*}{vegetation}             & \multirow{-2}{=}{bush, plants, tree} & tree trunk         & tree-trunk \\
\midrule
\rowcolor{rowshade} others, other\_flat  & --                  & GO                                  & -- \\
\midrule
sky (aux.)                               & sky                 & sky (aux.)                          & sky \\
\bottomrule
\end{tabularx}
\vspace{-0.2cm}
\end{table}

{\parskip=2pt
\noindent\textbf{Training and Evaluation.}
Both datasets provide a camera visibility mask that marks the voxels observed by the cameras.
For \approach, we only use the labels of the validation set of both datasets for evaluation.
Following previous work for occupancy prediction~\cite{tian2023occ3d, wang2024opus}, we only use 2 Hz frames as anchors for Occ3D-Waymo for training and train all models for the same number of epochs as for Occ3D-nuScenes due to the approximately similar resulting size.
For evaluation, we use the 10 Hz annotations.
}

\subsection{Pseudo-Label Generation}

\noindent\textbf{3D Reconstruction.} For each sequence, we input all keyframes at 2 Hz of all views jointly into $\phi$, resulting in approximately $40\times6 = 240$ images for a single nuScenes scene.
We condition $\phi$ on the recorded poses and intrinsics of every frame.
On nuScenes, we crop the bottom part of the rear camera to remove the visible part of the ego-vehicle.
We discard pixels with low confidence $\Omega < 0.38$ for Pi3X and outside the validity mask of MapAnything.
Additionally, we remove all points segmented as sky and all points with a predicted depth above 50 m.

{\parskip=2pt
\noindent\textbf{Scale Correction.}
As mentioned in \Cref{subsubsec:recon}, we estimate a scale correction for both reconstruction models to cover their failure cases.
For scenes in which the ego-vehicle remains static or moves less than 10 m, we fit a plane with RANSAC to the points pseudo-labeled as driveable surface.
To retrieve the scale correction, we then compare the vertical offset of all cameras per timestamp to the fitted ground plane with the known mounting height and take the median over all frames.
For all other scenes, we compute a RANSAC similarity alignment with the recorded poses of the datasets.
If the resulting scale correction differs by more than $1.25$ from the one estimated through the ground plane, we rely on the latter.
This covers failure cases in which the predicted poses have degraded performance, while the depth maps remain plausible.
}

{\parskip=2pt
\noindent\textbf{Open-Vocabulary Segmentation.}
We provide the prompts that were used to retrieve 2D semantic information and instance tracks with SAM3 and SAM3.1 in \Cref{tab:sam3_prompts}.
Following prior work~\cite{boeder2025shelfocc}, we exclude the classes \textit{others} and \textit{other\_flat} for nuScenes due to their ambiguous definition.
Similarly, we exclude the class \textit{general object (GO)} for Waymo from the prompts.
}

{\parskip=2pt
\noindent\textbf{Long-Term Motion Segmentation.}
We use $v_{\text{AD}}=8$\,km/h, $v_{\text{FD}}=3$\,km/h, and $v_{\text{ped}}=2$\,km/h for pedestrians to account for their slower speeds than vehicles.
An object strongly exceeds a threshold if $\text{AD}_k>2\,\tau_{\text{AD},k}$ or $\text{FD}_k>1.55\,\tau_{\text{FD},k}$.
}

\begin{table}[t]
  \centering
  \scriptsize
  \caption{Mapping from the Waymo LiDAR semantic classes to the Occ3D-Waymo classes,
    used to build the 2D depth and semantic rendering targets of the self-supervised
    pre-training stage of PreWorld. Points labeled \textit{undefined} are discarded. GO has no
    corresponding LiDAR class and therefore never occurs in the rendering targets, similar to the open-vocabulary prompts. Free space is supervised through the rendered depth.}
  \label{tab:waymo_lidarseg_to_occ3d}
  \begin{tabular}{@{}rll@{}}
    \toprule
    ID & Occ3D-Waymo class & Waymo LiDAR semantic classes \\
    \midrule
    0  & GO                & -- \\
    1  & Vehicle           & car, truck, bus, other vehicle \\
    2  & Pedestrian        & pedestrian \\
    3  & Sign              & sign \\
    4  & Bicyclist         & bicyclist, motorcyclist \\
    5  & Traffic light     & traffic light \\
    6  & Pole              & pole \\
    7  & Construction cone & construction cone \\
    8  & Bicycle           & bicycle \\
    9  & Motorcycle        & motorcycle \\
    10 & Building          & building \\
    11 & Vegetation        & vegetation \\
    12 & Tree trunk        & tree trunk \\
    13 & Road              & road, lane marker, curb, other ground \\
    14 & Walkable          & walkable, sidewalk \\
    \midrule
    -- & \textit{ignored}  & undefined \\
    \bottomrule
  \end{tabular}
  \vspace{-0.2cm}
\end{table}

\subsection{Model Training}
\textbf{SparseWorld-TC.}
For nuScenes, we train SparseWorld-TC-Small and -Small*~\cite{du2025sparseworld} with our HRS, but keep all other parameters unchanged.
Therefore, we train both models for the main table for $70$ epochs with an image resolution of $256 \times 704$, and use seven past frames.
We reimplement the DINOv3~\cite{simeoni2026dinov} ConvNeXt~\cite{liu2022convnet} version SparseWorld-TC-Small* since the official repository only provides the config for SparseWorld-TC-Small.
Moreover, we create a dataloader and adapt the model to work with Occ3D-Waymo.
Following previously trained occupancy prediction models~\cite{tian2023occ3d}, we use an image resolution of $640 \times 960$ for Waymo.
For the ablations, we train SparseWorld-TC-Small on Occ3D-nuScenes for $20$ epochs.

{\parskip=2pt
\noindent\textbf{SparseWorld.}
Following the paper, we use ResNet-50, initialized with the same nuImages weights as SparseWorld-TC-Small, for SparseWorld~\cite{dang2026sparseworld}, replacing only the supervision with our HRS and adding our MGL.
The model takes as input the current and four past camera frames with an image resolution of $256 \times 704$.
We train it on Occ3D-nuScenes for $6$ epochs of pre-training and $48$ epochs of end-to-end training, as stated in the paper~\cite{dang2026sparseworld}.
}

{\parskip=2pt
\noindent\textbf{PreWorld.} The official implementation of PreWorld~\cite{li2025semi} only supports Occ3D-nuScenes. We adapt it to Occ3D-Waymo and keep the original architecture and training recipe wherever possible.
Similar to our model, we resize all five views to $640 \times 960$ and train only on 2~Hz annotations.
Following the implementation of PreWorld for nuScenes, the model takes as input two past frames, and we perform $8$ epochs of pre-training for the 3D-label-free variant.
We subsequently train the fully supervised model for $18$ epochs with the 3D semantic occupancy ground truth labels.
Both variants use a Swin-B backbone, initialized from the trained 3D object detection model BEVDet4D-Stereo on nuScenes, as in the original paper.
As for nuScenes, the model is supervised only through differentiable volume rendering during the pre-training stage.
Following the original implementation for nuScenes, we generate the supervision targets by projecting the LiDAR points, along with their semantic labels, into the 2D image planes.
We use all available frames with LiDAR semantic labels.
Since the LiDAR segmentation labels differ slightly from the Occ3D-Waymo label definitions, we convert them as documented in \Cref{tab:waymo_lidarseg_to_occ3d}.
As for nuScenes, rays are also rendered from auxiliary frames $\pm$1.5 s around the anchor. %
For forecasting, an anchor is used only if labeled frames exist at all six future steps, and rays are sampled with class-balanced weights.
As in the nuScenes implementation, the auxiliary trajectory head is trained on the six future ego waypoints, and forecast horizons are added progressively over the $8$ epochs as in the original schedule.
For the fully-supervised stage, we use the same 2 Hz frames as for SparseWorld-TC and \approach, except for anchors whose 3~s future extends beyond the end of the sequence, mirroring PreWorld on nuScenes.
}

\subsection{Voxelization}
\label{sec:supp_voxelization}
Given the combined semantic point cloud $\mathcal{P}_t$ ($\mathcal{P}_{\text{static}} \cup \mathcal{P}_{\text{dynamic}, t}$) at timestamp $t$ and a pre-defined voxel grid, we perform a majority voting among all points per voxel.
A voxel cell is only marked empty if the number of points inside it falls below a certain point count.
Following ShelfOcc~\cite{boeder2025shelfocc}, we prioritize minority classes (traffic cone) over frequent classes (road, sidewalk, etc.) and drop static points in voxels that are traversed in other views and timestamps $F$ times more often than they are confirmed.
Moreover, we reimplement and apply the steps \textit{pinhole and cavity filling}, \textit{conservative neighborhood coherence}, and \textit{background cleanup} from the occupancy refinement steps in FreeOcc~\cite{caunes2026freeocc}.
To better learn fine details of objects but still efficiently capture large stuff classes during downstream training, we produce our pseudo-labels at two different resolutions, a coarse one, $\mathbf{V}_{t, \text{coarse}}$, following the voxel size in Occ3D~\cite{tian2023occ3d} and a fine-grained one, $\mathbf{V}_{t, \text{fine}}$, with doubled resolution.
Complementary to this, we store a dynamic mask $\mathbf{M}_{t, \text{dynamic, coarse/fine}}$ of the same shape as $\mathbf{V}_{t, \text{coarse/fine}}$ that evaluates to $1$ whenever a single point from $\mathcal{P}_{\text{dynamic}, t}$ falls into a voxel and $0$ otherwise, and which we use to guide the training of our world model.

\section{Additional Quantitative Results}\label{sec:supp_quantitative}

\begin{table}[t]
\scriptsize
\centering
\caption{Ablation of different backbone initializations.\label{tab:abl_backbone}}
\setlength{\tabcolsep}{5pt}
\begin{tabular}{cc|ccc} 
    \toprule
    Backbone & Init. Weights & Avg. mIoU $\uparrow$ & Avg. IoU $\uparrow$\\
    \midrule
    ResNet-50 & ImageNet-1K & 13.76 & 40.08\\
    \rowcolor{bannergray}
    ResNet-50 & nuImages & 14.04 & 40.19\\
    \midrule
    ConvNeXt-Base & ImageNet-1K & 13.95 & 40.75\\
    \rowcolor{bannergray}
    ConvNeXt-Base & DINOv3 & 14.92 & 42.71\\
    \bottomrule
\end{tabular}
\end{table}

\subsection{Image Backbone Initialization}
We study the effect of different backbone initializations of the underlying world model~\cite{du2025sparseworld} in \Cref{tab:abl_backbone}.
When using ResNet-50~\cite{he2016deep}, \approach achieves slightly improved results with initialization from nuImages (as used in the official implementation of SparseWorld-TC) compared to ImageNet-1K.
We therefore use the former for all other experiments.
ConvNeXt-Base with ImageNet-1K initialization as backbone performs better than ResNet-50 with the same initialization but slightly worse than ResNet-50 with nuImages initialization in terms of mIoU.
Leveraging self-supervised DINOv3 weights for ConvNeXt-Base (equal to \approach + SparseWorld-TC-Small* trained for 20 epochs) improves over it by almost one mIoU point.

\subsection{Ablation Study of LTMS Components}
We ablate additional design decisions for LTMS on the pseudo-label quality in \Cref{tab:abl_motion_smoothing}.
We observe that applying Kalman filter smoothing on the BEV trajectories slightly improves mIoU and RayIoU.
Using the AD/FD overrides for cases in which AD and FD decisions largely disagree provides a larger benefit for both metrics.

\begin{table}[t]
    \scriptsize
    \centering
    \caption{Ablation study for LTMS design decisions.\label{tab:abl_motion_smoothing}}
    \begin{tabular}{cc|cc} 
        \toprule
        Smoothing & AD/FD Override & mIoU $\uparrow$ & RayIoU $\uparrow$\\
        \midrule
        \xmark & \cmark & 22.87 & 21.88\\ %
        \cmark & \xmark & 22.78 & 21.67\\
        \rowcolor{bannergray}
        \cmark & \cmark & \textbf{22.94} & \textbf{22.03}\\
        \bottomrule
    \end{tabular}
    \vspace{-0.2cm}
\end{table}

\subsection{Time Analysis}
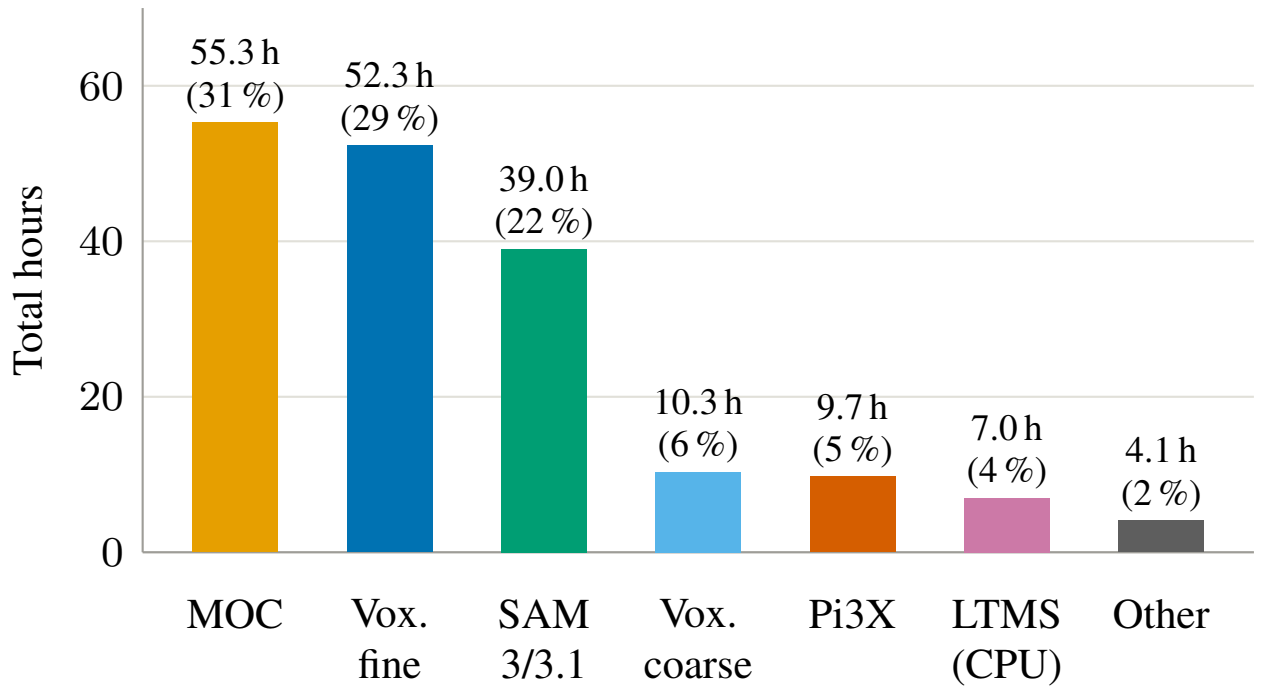
\begin{figure}[t]
  \centering
  \definecolor{pwtSamThree}{HTML}{009E73}%
\definecolor{pwtPiThree}{HTML}{D55E00}%
\definecolor{pwtLtms}{HTML}{CC79A7}%
\definecolor{pwtSamThreeD}{HTML}{E69F00}%
\definecolor{pwtVoxCoarse}{HTML}{56B4E9}%
\definecolor{pwtVoxFine}{HTML}{0072B2}%
\definecolor{pwtOther}{HTML}{5E5E5E}%
\definecolor{pwtInk}{HTML}{52514E}%
\definecolor{pwtGrid}{HTML}{E1E0D9}%
\begin{tikzpicture}
\begin{axis}[
    ybar, bar shift=0pt, bar width=16pt,
    scale only axis, width=\linewidth-28pt, height=3.6cm,  %
    xmin=-0.6, xmax=6.6,
    xtick={0,...,6},
    xticklabels={{MOC}, {Vox.\\fine}, {SAM\\3/3.1}, {Vox.\\coarse}, {Pi3X}, {LTMS\\(CPU)}, {Other}},
    xticklabel style={align=center, anchor=north, yshift=-1pt},
    ymin=0, ymax=70, ytick distance=20,
    ylabel={Total hours},
    font=\footnotesize, tick label style={font=\footnotesize, text=black}, label style={text=black},
    axis x line*=bottom, axis y line*=left, tick style={draw=none},
    ymajorgrids, grid style={pwtGrid}, axis line style={pwtGrid!70!black},
    clip=false,
]
\addplot[fill=pwtSamThreeD, draw=none] coordinates {(0,55.29)};
\addplot[fill=pwtVoxFine, draw=none] coordinates {(1,52.32)};
\addplot[fill=pwtSamThree, draw=none] coordinates {(2,39.04)};
\addplot[fill=pwtVoxCoarse, draw=none] coordinates {(3,10.34)};
\addplot[fill=pwtPiThree, draw=none] coordinates {(4,9.72)};
\addplot[fill=pwtLtms, draw=none] coordinates {(5,7.00)};
\addplot[fill=pwtOther, draw=none] coordinates {(6,4.10)};
\foreach \x/\h/\lab/\pct in {0/55.29/55.3/31, 1/52.32/52.3/29, 2/39.04/39.0/22, 3/10.34/10.3/6, 4/9.72/9.7/5, 5/7.00/7.0/4, 6/4.10/4.1/2}{
  \edef\tmp{\noexpand\node[anchor=south, align=center, font=\noexpand\scriptsize, text=black, inner sep=1.5pt]
    at (axis cs:\x,\h) {\lab\noexpand\,h\noexpand\\(\pct\noexpand\,\noexpand\%)};}\tmp
}
\end{axis}
\end{tikzpicture}%
  \caption{Total time per pseudo-labeling component for the 150 nuScenes validation scenes. LTMS runs on the CPU, in parallel to the GPU processes.}\label{fig:time_total}
\end{figure}

\begin{figure}[t]
  \centering
  \definecolor{pwtSamThree}{HTML}{009E73}%
\definecolor{pwtPiThree}{HTML}{D55E00}%
\definecolor{pwtLtms}{HTML}{CC79A7}%
\definecolor{pwtSamThreeD}{HTML}{E69F00}%
\definecolor{pwtVoxCoarse}{HTML}{56B4E9}%
\definecolor{pwtVoxFine}{HTML}{0072B2}%
\definecolor{pwtOther}{HTML}{5E5E5E}%
\definecolor{pwtInk}{HTML}{52514E}%
\definecolor{pwtGrid}{HTML}{E1E0D9}%
\pgfplotstableread{
idx sam3 pi3x dynamics sam3d vox04 vox02 other
0 13.42 3.36 2.36 1.38 3.43 11.97 1.16
1 14.10 5.42 2.93 3.62 2.87 9.85 1.41
2 11.37 3.87 2.43 2.97 3.67 16.36 1.42
3 12.46 3.28 2.46 3.54 3.64 15.81 1.48
4 11.93 4.38 2.55 4.64 3.50 16.06 1.36
5 11.77 3.11 2.59 4.94 3.73 18.77 1.39
6 13.60 7.43 2.76 5.12 3.05 13.65 1.39
7 13.87 3.93 2.47 5.66 3.61 16.21 1.37
8 13.58 3.12 2.55 8.36 3.53 14.00 2.20
9 12.83 3.28 2.37 7.52 3.77 16.72 2.20
10 12.07 3.19 2.47 4.50 4.21 21.82 1.58
11 13.66 3.21 2.55 6.15 3.91 19.17 1.38
12 13.68 4.14 2.76 10.34 3.25 15.42 1.25
13 14.88 4.18 2.72 6.92 3.88 17.82 1.37
14 13.06 3.17 2.49 8.90 3.81 19.11 1.40
15 15.04 4.56 3.31 6.25 3.51 18.02 1.42
16 12.77 3.37 2.32 7.18 4.18 21.47 1.13
17 15.89 3.26 2.69 7.29 3.75 17.76 2.28
18 12.86 3.44 2.58 8.41 4.06 20.92 1.41
19 11.93 3.40 2.52 6.55 4.66 23.29 1.46
20 13.20 3.27 2.46 7.04 4.31 22.15 1.40
21 14.07 4.36 2.39 9.69 3.86 17.68 2.12
22 13.62 3.35 2.56 5.55 4.68 22.90 1.58
23 12.34 3.27 2.62 9.86 3.86 21.23 1.43
24 15.51 3.63 2.62 9.65 4.07 17.57 1.68
25 12.77 3.12 2.52 10.92 4.12 20.73 1.41
26 14.30 4.46 2.78 7.21 4.08 21.40 1.42
27 12.65 3.22 2.59 9.07 4.30 22.49 1.42
28 15.64 3.30 2.77 10.01 4.01 18.74 1.59
29 15.30 4.36 3.30 13.13 4.30 14.15 1.60
30 11.76 3.24 2.47 9.76 4.33 23.04 1.61
31 12.74 3.31 2.48 6.79 4.55 24.34 2.58
32 12.14 3.33 2.54 12.15 3.92 21.50 1.41
33 15.78 3.69 2.62 9.85 4.02 20.27 1.30
34 14.26 3.38 2.77 10.86 3.85 21.15 1.40
35 15.90 3.56 2.85 10.44 3.87 19.94 1.30
36 14.32 3.54 2.40 12.45 3.89 20.48 1.13
37 14.68 6.96 2.99 14.61 3.77 13.77 1.51
38 17.31 3.30 3.05 18.20 3.99 11.94 1.43
39 14.62 5.67 2.79 10.84 3.87 20.16 2.41
40 14.18 3.27 2.63 15.79 3.84 19.40 1.33
41 14.21 3.41 2.48 12.85 4.14 21.42 2.02
42 14.52 3.54 2.59 13.18 4.00 21.33 1.45
43 14.36 3.83 2.79 7.00 5.05 25.77 1.90
44 12.94 3.12 2.65 16.15 3.66 20.79 1.49
45 13.61 3.34 2.63 15.54 3.92 20.78 1.38
46 14.68 4.06 3.14 14.64 4.06 19.25 1.60
47 20.04 3.23 2.61 8.63 4.12 21.55 2.62
48 14.00 3.57 2.46 12.85 4.44 23.45 2.20
49 14.64 3.68 2.61 15.97 4.01 20.98 1.30
50 12.14 3.93 2.60 14.01 4.70 24.23 1.90
51 17.65 4.05 2.99 12.89 3.95 20.60 1.44
52 16.70 3.36 2.85 17.36 3.87 18.55 1.55
53 12.19 3.16 2.59 18.09 4.21 22.61 1.50
54 14.84 5.96 2.87 14.16 4.37 20.92 1.53
55 15.03 3.64 2.65 13.46 4.84 24.68 1.45
56 16.14 3.35 2.66 19.41 3.62 19.69 1.27
57 15.70 3.39 2.87 17.28 4.08 20.80 2.34
58 15.23 3.43 2.68 18.08 3.93 21.00 2.11
59 13.09 3.06 2.64 21.00 3.93 20.65 2.26
60 14.53 3.60 2.65 18.62 3.99 21.95 1.51
61 15.78 5.06 2.94 14.57 4.29 22.78 1.61
62 16.20 3.24 2.77 18.01 3.88 20.58 2.40
63 13.92 3.30 2.48 18.44 4.15 22.65 2.19
64 13.60 8.55 2.64 13.19 4.69 23.51 1.64
65 17.14 3.80 2.70 13.89 4.87 24.60 1.69
66 16.55 3.30 2.74 22.05 3.60 19.33 1.44
67 16.86 3.56 2.75 19.88 4.30 21.55 1.41
68 15.19 3.32 2.58 20.64 4.44 23.10 1.29
69 14.74 7.10 2.42 17.53 4.22 23.61 1.24
70 15.72 4.35 2.82 20.80 4.20 22.17 1.44
71 16.62 3.33 2.54 22.56 3.99 20.79 2.11
72 17.66 8.02 2.96 13.77 4.29 22.99 2.36
73 16.80 6.52 3.00 18.33 4.16 21.63 1.61
74 16.40 3.25 2.78 22.33 4.11 21.92 1.52
75 17.15 3.71 2.81 19.62 4.50 23.04 1.50
76 15.38 3.24 2.71 25.60 3.80 20.56 1.39
77 17.76 3.57 3.19 21.54 4.26 20.97 1.52
78 13.89 3.14 2.65 22.02 4.68 23.94 2.52
79 16.38 3.22 2.75 21.47 4.55 23.14 1.45
80 16.80 3.30 2.70 19.00 5.02 23.73 2.53
81 16.43 3.79 2.76 23.64 3.80 20.59 2.24
82 16.70 3.23 3.25 23.52 4.13 21.05 1.44
83 15.90 6.01 2.89 23.71 3.63 19.97 1.40
84 17.01 3.38 2.87 24.62 4.03 20.49 1.52
85 19.00 3.39 3.24 22.45 3.66 20.83 1.37
86 16.68 3.28 3.01 25.45 4.02 20.90 1.48
87 15.60 3.15 2.98 25.95 4.00 20.75 2.43
88 17.58 7.23 2.95 21.21 4.11 20.95 1.45
89 15.85 3.37 3.00 27.22 4.63 19.99 1.55
90 17.33 3.67 2.82 25.20 4.16 21.21 1.46
91 15.78 3.40 2.40 26.28 4.37 21.52 2.18
92 15.05 4.26 2.88 26.59 4.12 22.04 1.31
93 16.53 3.92 2.94 25.97 4.04 21.65 1.50
94 16.05 3.57 2.73 28.11 3.97 20.79 1.39
95 16.12 3.15 2.73 23.47 4.75 23.88 2.60
96 16.90 3.32 2.90 26.73 4.04 21.74 1.44
97 15.03 5.92 2.61 26.87 4.21 21.24 1.20
98 15.80 3.37 2.69 22.92 4.94 25.89 1.60
99 15.64 3.57 2.76 26.77 4.47 22.45 1.56
100 16.46 3.24 2.77 28.34 3.97 20.45 2.21
101 16.53 3.37 2.75 29.24 4.16 20.42 1.38
102 23.62 3.34 3.40 21.69 4.08 21.00 1.36
103 16.66 7.01 2.87 24.81 4.13 21.57 1.45
104 18.81 3.47 2.98 25.87 4.19 21.70 1.59
105 15.26 4.18 2.68 30.33 3.92 21.05 1.32
106 17.20 3.38 2.69 27.79 4.14 21.41 2.22
107 17.32 3.31 2.84 28.04 4.31 21.61 1.51
108 15.05 3.66 2.77 30.71 4.26 21.63 1.34
109 16.98 3.44 2.64 24.74 4.86 26.70 1.57
110 14.56 5.32 2.66 30.09 4.42 22.75 1.42
111 14.97 4.85 2.38 29.55 4.50 23.86 1.29
112 16.46 3.27 2.98 29.44 4.47 23.32 1.47
113 16.07 4.77 2.83 30.50 4.13 20.89 2.25
114 16.74 3.47 2.85 32.50 4.01 20.81 1.46
115 18.13 3.23 2.84 31.57 3.91 20.60 2.22
116 18.64 3.74 3.00 28.46 4.58 22.75 1.60
117 19.40 4.21 3.15 29.73 3.98 20.96 1.36
118 14.91 3.23 2.77 33.55 4.32 22.78 1.40
119 16.20 3.28 2.89 35.29 3.83 20.78 1.15
120 17.79 4.58 3.43 28.58 4.40 23.19 1.56
121 18.02 3.79 3.00 32.48 3.97 21.01 1.42
122 14.00 3.17 2.70 37.57 4.09 20.91 1.46
123 16.89 3.92 2.97 30.30 4.84 23.45 1.64
124 17.26 3.83 3.12 29.44 4.72 24.34 1.72
125 15.35 4.10 2.88 34.94 4.41 22.70 1.32
126 15.83 3.32 3.06 38.61 3.69 20.13 1.39
127 17.40 3.63 2.82 29.09 5.18 26.92 1.68
128 16.77 3.30 2.62 35.22 4.75 23.80 1.27
129 17.64 4.39 3.12 35.88 3.87 21.03 2.17
130 19.63 3.96 3.23 32.05 4.73 24.41 1.65
131 15.56 3.39 3.11 40.65 3.95 20.68 2.59
132 18.59 4.19 3.30 37.95 3.84 20.85 1.51
133 15.49 3.86 3.16 42.13 3.82 20.47 1.51
134 17.30 3.40 3.03 40.72 3.81 21.28 1.44
135 16.34 3.37 2.81 42.43 4.37 21.16 1.42
136 17.27 6.05 2.98 38.41 4.15 21.79 1.43
137 16.94 4.72 3.06 38.87 4.53 22.90 1.32
138 18.14 3.49 3.31 42.17 4.05 21.38 1.57
139 15.43 3.41 2.86 47.32 3.82 21.41 1.46
140 18.87 4.19 2.86 39.54 4.79 23.66 2.48
141 18.05 3.47 2.87 44.22 4.48 22.52 1.52
142 16.44 3.29 3.52 49.55 4.99 16.92 2.55
143 17.21 5.06 3.22 47.45 4.44 18.62 1.61
144 16.34 3.26 2.98 48.62 4.08 21.01 1.53
145 19.20 3.32 3.26 48.48 3.62 19.23 1.34
146 18.41 3.39 3.05 47.36 4.13 21.61 2.40
147 17.59 3.48 3.18 49.89 4.38 23.37 2.47
148 17.16 3.32 3.15 53.79 4.80 24.01 1.68
149 15.67 3.48 3.03 64.03 4.12 22.62 1.48
}\pwtPerSceneTable%
\begin{tikzpicture}
\begin{axis}[
    ybar stacked, bar width=0.8,
    scale only axis, width=\linewidth-34pt, height=3.6cm,  %
    xmin=-0.5, xmax=149.5, xtick=\empty,
    ymin=0, ymax=120, ytick distance=30,
    xlabel={150 scenes, sorted by total time},
    ylabel={Minutes per scene},
    font=\footnotesize, tick label style={font=\footnotesize, text=black}, label style={text=black},
    axis x line*=bottom, axis y line*=left, tick style={draw=none},
    ymajorgrids, grid style={pwtGrid}, axis line style={pwtGrid!70!black},
    legend style={at={(0.5,1.02)}, anchor=south, xshift=-16pt,  %
                  draw=none, fill=none, font=\scriptsize, text=black,
                  inner sep=0pt, row sep=1pt,
                  /tikz/every even column/.append style={column sep=5pt}},
    legend columns=4, legend cell align=left, reverse legend,
    legend image code/.code={\fill[#1] (0pt,-2.5pt) rectangle (7pt,3.5pt);},
    clip=false,
]
\addplot[fill=pwtOther, draw=none] table[x=idx, y=other] {\pwtPerSceneTable};
\addplot[fill=pwtLtms, draw=none] table[x=idx, y=dynamics] {\pwtPerSceneTable};
\addplot[fill=pwtPiThree, draw=none] table[x=idx, y=pi3x] {\pwtPerSceneTable};
\addplot[fill=pwtVoxCoarse, draw=none] table[x=idx, y=vox04] {\pwtPerSceneTable};
\addplot[fill=pwtSamThree, draw=none] table[x=idx, y=sam3] {\pwtPerSceneTable};
\addplot[fill=pwtVoxFine, draw=none] table[x=idx, y=vox02] {\pwtPerSceneTable};
\addplot[fill=pwtSamThreeD, draw=none] table[x=idx, y=sam3d] {\pwtPerSceneTable};
\legend{{Other}, {LTMS (CPU)}, {Pi3X}, {Vox.\ coarse}, {SAM~3/3.1}, {Vox.\ fine}, {MOC}}
\coordinate (medL) at (axis cs:-0.5,72.32);
\coordinate (medR) at (axis cs:149.5,72.32);
\coordinate (medT) at (axis cs:0,72.32);
\end{axis}
\draw[pwtInk, dashed, line width=0.5pt] (medL) -- (medR);
\node[anchor=south west, font=\scriptsize, text=black, inner sep=1.5pt] at (medT) {median 72\,min};
\end{tikzpicture}%
  \caption{Pseudo-labeling time per scene of the nuScenes validation set, divided into components. The median is 72 minutes.}\label{fig:time_scene}
  \vspace{-0.2cm}
\end{figure}
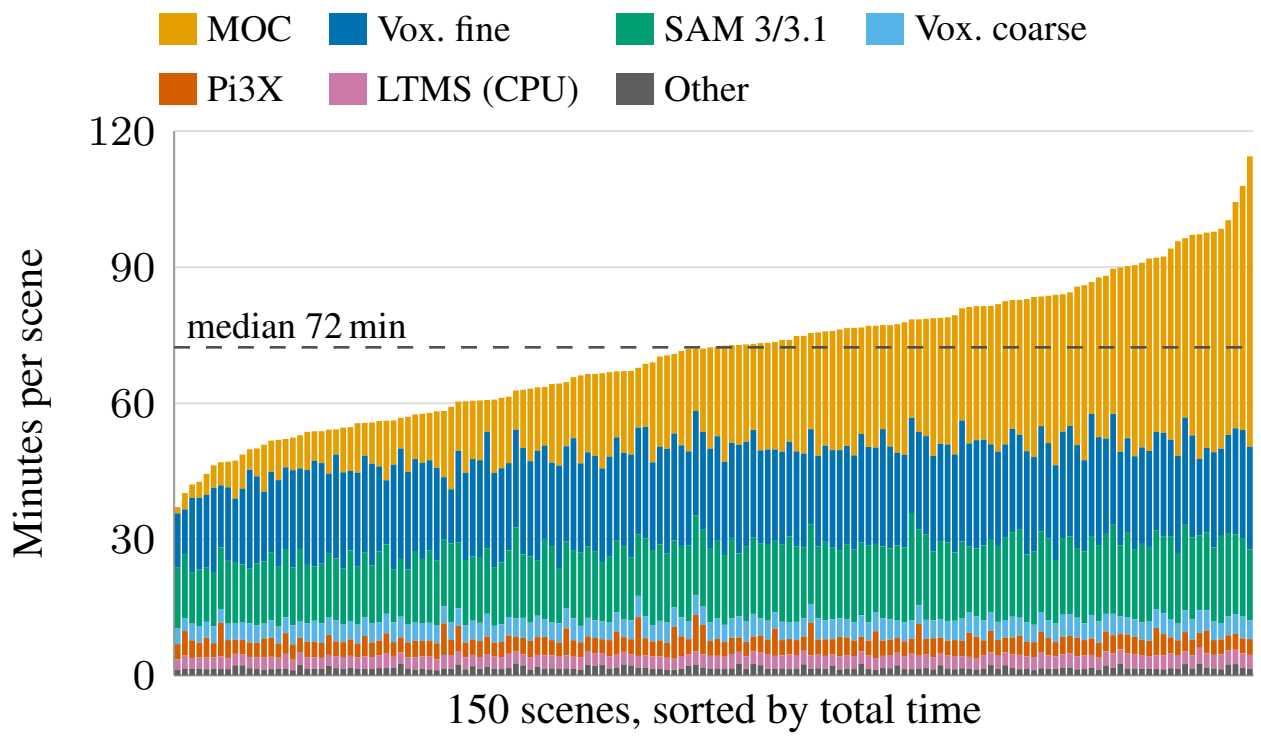

We provide a time analysis of our motion-aware pseudo-labeling stage, divided into the time spent per component in \Cref{fig:time_total} and per scene in \Cref{fig:time_scene}.
For this analysis, we use a machine with two AMD EPYC 7452 CPUs with 256 GB RAM and six RTX A6000 GPUs.
We conduct measurements on the 150 nuScenes validation sequences.
It should be emphasized that the pseudo-labels for the validation set were not used to train any model.

Processing all 150 scenes takes about 171 GPU hours plus 7 CPU hours, with a median of 72\,min per scene.
Extrapolating to the 700 training scenes yields roughly 800 GPU hours.
MOC consumes the most time due to the inference cost of SAM 3D Objects that grows linearly with the number of moving objects per scene.
Generating the fine voxel grid requires about the same total time but is roughly stable per scene.
SAM3 and SAM3.1 inference is the third-largest cost and increases only slightly with the number of objects per scene.
Coarse voxelization, Pi3X inference, and other minor steps, such as I/O and scale alignment, require fewer GPU hours, while our LTMS is able to run in parallel with voxelization and MOC on the CPU.

\begin{table*}[t]
\centering
\scriptsize
\caption{Per-class 4D occupancy forecasting performance on Occ3D-nuScenes in mIoU. $\dagger$ means results reproduced by us with the released checkpoint. mIoU is the mean IoU over all 17 classes, while mIoU* excludes \textit{others} and \textit{other\_flat}. SW-TC represents SparseWorld-TC, and S Small.}
\label{tab:per_class_forecasting}
\setlength{\tabcolsep}{2pt}
\begin{tabular}{cl|ccc|*{17}{c}}
\toprule
 & Method & mIoU & mIoU* & IoU &
\colhead{cOthers}{others} &
\colhead{cBarrier}{barrier} &
\colhead{cBicycle}{bicycle} &
\colhead{cBus}{bus} &
\colhead{cCar}{car} &
\colhead{cConsVeh}{cons. vehicle} &
\colhead{cMotorcycle}{motorcycle} &
\colhead{cPedestrian}{pedestrian} &
\colhead{cTrafficCone}{traffic cone} &
\colhead{cTrailer}{trailer} &
\colhead{cTruck}{truck} &
\colhead{cDrivSurf}{driv. surf.} &
\colhead{cOtherFlat}{other flat} &
\colhead{cSidewalk}{sidewalk} &
\colhead{cTerrain}{terrain} &
\colhead{cManmade}{manmade} &
\colhead{cVegetation}{vegetation} \\
\midrule
\multirow{5.5}{*}{$t{+}1$\,s}
 & \multicolumn{21}{l}{\textit{Fully Supervised}} \\
 & SparseWorld-TC-S~\cite{du2025sparseworld}$\dagger$ & 27.81 & 28.16 & 50.24 & 9.73 & 35.82 & 11.06 & 25.30 & 30.61 & 19.79 & 9.65 & 6.42 & 14.92 & 20.33 & 23.30 & 73.48 & 40.74 & 45.31 & 43.53 & 31.52 & 31.29 \\
\cmidrule{2-22}
 & \multicolumn{21}{l}{\textit{3D Label-Free}} \\
\rowcolor{bannergray}
\cellcolor{white} & \approach (+ SW-TC-S) & 15.72 & 17.82 & 43.17 & 0.00 & 15.23 & \textbf{1.07} & \textbf{16.26} & \textbf{20.31} & 9.32 & 1.58 & 2.72 & 3.96 & \textbf{2.51} & \textbf{15.26} & \textbf{63.50} & 0.00 & 37.94 & 34.12 & 21.68 & 21.80 \\
\rowcolor{bannergray}
\cellcolor{white} & \approach (+ SW-TC-S*) & \textbf{15.97} & \textbf{18.10} & \textbf{44.53} & 0.00 & \textbf{15.49} & 0.98 & 14.53 & 18.83 & \textbf{10.74} & \textbf{1.71} & \textbf{3.33} & \textbf{5.76} & 2.01 & 15.10 & 63.05 & 0.00 & \textbf{39.38} & \textbf{35.16} & \textbf{23.07} & \textbf{22.31} \\
\midrule
\multirow{5.5}{*}{$t{+}2$\,s}
 & \multicolumn{21}{l}{\textit{Fully Supervised}} \\
 & SparseWorld-TC-S~\cite{du2025sparseworld}$\dagger$ & 25.31 & 25.45 & 48.82 & 9.02 & 33.53 & 8.30 & 18.30 & 24.44 & 17.89 & 7.60 & 3.56 & 12.16 & 17.88 & 19.12 & 72.45 & 39.63 & 44.18 & 42.42 & 30.02 & 29.83 \\
\cmidrule{2-22}
 & \multicolumn{21}{l}{\textit{3D Label-Free}} \\
\rowcolor{bannergray}
\cellcolor{white} & \approach (+ SW-TC-S) & 14.73 & 16.69 & 42.46 & 0.00 & \textbf{14.36} & \textbf{0.84} & \textbf{11.56} & \textbf{17.40} & 8.40 & 1.20 & 1.63 & 3.31 & \textbf{3.07} & 12.90 & \textbf{62.88} & 0.00 & 37.18 & 33.46 & 21.12 & 21.08 \\
\rowcolor{bannergray}
\cellcolor{white} & \approach (+ SW-TC-S*) & \textbf{15.04} & \textbf{17.05} & \textbf{43.83} & 0.00 & 14.07 & 0.82 & 11.19 & 16.47 & \textbf{10.13} & \textbf{1.29} & \textbf{1.89} & \textbf{4.79} & 2.14 & \textbf{13.43} & 62.67 & 0.00 & \textbf{38.66} & \textbf{34.44} & \textbf{22.23} & \textbf{21.47} \\
\midrule
\multirow{5.5}{*}{$t{+}3$\,s}
 & \multicolumn{21}{l}{\textit{Fully Supervised}} \\
 & SparseWorld-TC-S~\cite{du2025sparseworld}$\dagger$ & 23.10 & 23.10 & 47.01 & 8.35 & 30.69 & 5.65 & 13.81 & 20.50 & 16.05 & 6.07 & 2.11 & 9.38 & 15.57 & 16.27 & 70.87 & 37.82 & 42.50 & 40.78 & 28.20 & 28.02 \\
\cmidrule{2-22}
 & \multicolumn{21}{l}{\textit{3D Label-Free}} \\
\rowcolor{bannergray}
\cellcolor{white} & \approach (+ SW-TC-S) & 13.98 & 15.84 & 41.48 & 0.00 & \textbf{13.39} & 0.71 & 8.89 & \textbf{15.51} & 7.51 & 0.93 & 1.23 & 2.43 & \textbf{5.26} & 10.86 & \textbf{61.97} & 0.00 & 36.01 & 32.42 & 20.40 & 20.12 \\
\rowcolor{bannergray}
\cellcolor{white} & \approach (+ SW-TC-S*) & \textbf{14.29} & \textbf{16.19} & \textbf{42.78} & 0.00 & 12.78 & \textbf{0.73} & \textbf{9.44} & 15.01 & \textbf{9.30} & \textbf{1.06} & \textbf{1.31} & \textbf{3.75} & 2.48 & \textbf{12.13} & 61.96 & 0.00 & \textbf{37.60} & \textbf{33.42} & \textbf{21.35} & \textbf{20.58} \\
\bottomrule
\end{tabular}
\end{table*}

\begin{table*}[t]
\centering
\scriptsize
\caption{Per-class 4D occupancy forecasting performance on Occ3D-nuScenes in RayIoU~\cite{liu2024fully}. $\dagger$ means results reproduced by us with the released checkpoint. RayIoU is the mean over all 17 classes and the distance thresholds of 1, 2, and 4 m, and RayIoU@$d$ is the mean over the classes at threshold $d$. The per-class values are averaged over the three thresholds. SW-TC represents SparseWorld-TC, and S Small.}
\label{tab:per_class_forecasting_rayiou}
\setlength{\tabcolsep}{1.25pt}
\begin{tabular}{cl|cccc|*{17}{c}}
\toprule
 & Method & RayIoU & @1\,m & @2\,m & @4\,m &
\colhead{cOthers}{others} &
\colhead{cBarrier}{barrier} &
\colhead{cBicycle}{bicycle} &
\colhead{cBus}{bus} &
\colhead{cCar}{car} &
\colhead{cConsVeh}{cons. vehicle} &
\colhead{cMotorcycle}{motorcycle} &
\colhead{cPedestrian}{pedestrian} &
\colhead{cTrafficCone}{traffic cone} &
\colhead{cTrailer}{trailer} &
\colhead{cTruck}{truck} &
\colhead{cDrivSurf}{driv. surf.} &
\colhead{cOtherFlat}{other flat} &
\colhead{cSidewalk}{sidewalk} &
\colhead{cTerrain}{terrain} &
\colhead{cManmade}{manmade} &
\colhead{cVegetation}{vegetation} \\
\midrule
\multirow{5.5}{*}{$t{+}1$\,s}
 & \multicolumn{22}{l}{\textit{Fully Supervised}} \\
 & SparseWorld-TC-S~\cite{du2025sparseworld}$\dagger$ & 31.79 & 25.91 & 32.26 & 37.21 & 8.40 & 40.24 & 14.31 & 42.16 & 42.98 & 26.19 & 10.15 & 16.83 & 24.48 & 22.75 & 42.35 & 65.42 & 35.24 & 31.87 & 34.27 & 44.62 & 38.25 \\
\cmidrule{2-23}
 & \multicolumn{22}{l}{\textit{3D Label-Free}} \\
\rowcolor{bannergray}
\cellcolor{white} & \approach (+ SW-TC-S) & 17.31 & 12.39 & 17.56 & 21.98 & 0.00 & 25.63 & \textbf{2.05} & \textbf{25.74} & \textbf{27.78} & \textbf{17.87} & \textbf{2.73} & 6.96 & 5.69 & \textbf{2.42} & \textbf{32.60} & 47.90 & 0.00 & 16.51 & 20.74 & 32.69 & 26.95 \\
\rowcolor{bannergray}
\cellcolor{white} & \approach (+ SW-TC-S*) & \textbf{17.63} & \textbf{12.61} & \textbf{17.99} & \textbf{22.28} & 0.00 & \textbf{26.66} & 1.85 & 23.95 & 26.98 & 17.38 & 1.67 & \textbf{8.90} & \textbf{8.56} & 2.02 & 30.82 & \textbf{48.73} & 0.00 & \textbf{17.30} & \textbf{22.10} & \textbf{34.59} & \textbf{28.18} \\
\midrule
\multirow{5.5}{*}{$t{+}2$\,s}
 & \multicolumn{22}{l}{\textit{Fully Supervised}} \\
 & SparseWorld-TC-S~\cite{du2025sparseworld}$\dagger$ & 28.23 & 22.82 & 28.57 & 33.30 & 7.85 & 38.05 & 10.88 & 31.53 & 34.74 & 24.88 & 7.95 & 9.49 & 21.53 & 18.29 & 37.33 & 62.43 & 33.13 & 30.11 & 32.62 & 42.66 & 36.39 \\
\cmidrule{2-23}
 & \multicolumn{22}{l}{\textit{3D Label-Free}} \\
\rowcolor{bannergray}
\cellcolor{white} & \approach (+ SW-TC-S) & 16.04 & 11.35 & 16.17 & 20.59 & 0.00 & 24.48 & 1.10 & \textbf{20.24} & \textbf{25.17} & \textbf{16.73} & \textbf{2.16} & 3.55 & 5.12 & \textbf{4.23} & \textbf{29.76} & 46.10 & 0.00 & 15.92 & 20.10 & 31.90 & 26.07 \\
\rowcolor{bannergray}
\cellcolor{white} & \approach (+ SW-TC-S*) & \textbf{16.38} & \textbf{11.68} & \textbf{16.66} & \textbf{20.79} & 0.00 & \textbf{25.41} & \textbf{1.57} & 19.55 & 24.37 & 16.22 & 1.43 & \textbf{4.98} & \textbf{7.77} & 2.22 & 28.46 & \textbf{47.28} & 0.00 & \textbf{16.78} & \textbf{21.39} & \textbf{33.72} & \textbf{27.30} \\
\midrule
\multirow{5.5}{*}{$t{+}3$\,s}
 & \multicolumn{22}{l}{\textit{Fully Supervised}} \\
 & SparseWorld-TC-S~\cite{du2025sparseworld}$\dagger$ & 25.40 & 20.38 & 25.64 & 30.19 & 7.51 & 35.62 & 7.61 & 25.09 & 29.61 & 23.62 & 6.68 & 5.87 & 18.02 & 13.96 & 33.87 & 59.33 & 31.19 & 28.27 & 30.78 & 40.51 & 34.31 \\
\cmidrule{2-23}
 & \multicolumn{22}{l}{\textit{3D Label-Free}} \\
\rowcolor{bannergray}
\cellcolor{white} & \approach (+ SW-TC-S) & 14.98 & 10.53 & 15.08 & 19.34 & 0.00 & 23.10 & 0.95 & 16.43 & \textbf{23.04} & \textbf{15.75} & \textbf{1.69} & 2.43 & 4.25 & \textbf{5.29} & 26.63 & 44.64 & 0.00 & 15.25 & 19.41 & 30.80 & 25.06 \\
\rowcolor{bannergray}
\cellcolor{white} & \approach (+ SW-TC-S*) & \textbf{15.44} & \textbf{10.96} & \textbf{15.69} & \textbf{19.66} & 0.00 & \textbf{23.96} & \textbf{1.57} & \textbf{16.77} & 22.41 & 15.70 & 1.13 & \textbf{3.20} & \textbf{6.76} & 2.21 & \textbf{26.79} & \textbf{46.09} & 0.00 & \textbf{16.20} & \textbf{20.63} & \textbf{32.70} & \textbf{26.32} \\
\bottomrule
\end{tabular}
\vspace{-0.2cm}
\end{table*}

\begin{table*}[t]
\centering
\scriptsize
\caption{Per-class 3D semantic occupancy estimation performance on Occ3D-nuScenes. mIoU excludes the \textit{others} and \textit{other flat} classes.}
\label{tab:occ3d_results}
\setlength{\tabcolsep}{3pt}
\begin{tabular}{l|cc|*{15}{c}}
\toprule
Method & mIoU & IoU &
\colhead{cBarrier}{barrier} &
\colhead{cBicycle}{bicycle} &
\colhead{cBus}{bus} &
\colhead{cCar}{car} &
\colhead{cConsVeh}{cons. vehicle} &
\colhead{cMotorcycle}{motorcycle} &
\colhead{cPedestrian}{pedestrian} &
\colhead{cTrafficCone}{traffic cone} &
\colhead{cTrailer}{trailer} &
\colhead{cTruck}{truck} &
\colhead{cDrivSurf}{driv. surf.} &
\colhead{cSidewalk}{sidewalk} &
\colhead{cTerrain}{terrain} &
\colhead{cManmade}{manmade} &
\colhead{cVegetation}{vegetation} \\
\midrule
SelfOcc~\cite{huang2024selfocc}          & 10.54 & 44.05 & 0.15  & 0.66  & 5.46  & 12.54 & 0.00 & 0.80  & 2.10  & 0.00  & 0.00 & 8.25  & 55.49 & 26.30 & 26.54 & 14.22 & 5.60 \\
OccNeRF~\cite{zhang2025occnerf}          & 10.81 & 46.43 & 0.83  & 0.82  & 5.13  & 12.49 & 3.50 & 0.23  & 3.10  & 1.84  & 0.52 & 3.90  & 52.62 & 20.81 & 24.75 & 18.45 & 13.19 \\
GaussianOcc~\cite{gan2025gaussianocc}  & 11.26 & 42.91 & 1.79  & 5.82  & 14.58 & 13.55 & 1.30 & 2.82  & 7.95  & 9.76  & 0.56 & 9.61  & 44.59 & 20.10 & 17.58 & 8.61  & 10.29 \\
GaussTR~\cite{jiang2025gausstr}          & 13.26 & 44.54 & 2.09  & 5.22  & 14.07 & 20.43 & 5.70 & 7.08  & 5.12  & 3.93  & 0.92 & 13.36 & 39.44 & 15.68 & 22.89 & 21.17 & 21.87 \\
EasyOcc~\cite{hayes2026easy3d}          & 15.96 & 38.86 & 1.85  & 8.18  & 16.66 & 22.12 & 1.01 & 7.74  & \underline{14.74} & 12.84 & 0.98 & 13.76 & 55.91 & 27.96 & 22.73 & 15.77 & 17.20 \\
TT-Occ-Camera~\cite{zhang2026test} & 16.70 & - & \textbf{21.51} & 10.46 & 10.70 & 14.72 & 11.89 & 12.33 & 9.71 & 12.20 & 4.37 & 7.92 & 48.27 & 23.70 & 28.31 & 14.13 & 20.24\\
GaussianFlowOcc~\cite{boeder2025gaussianflowocc} & 17.08 & 46.91 & 6.75  & 9.68  & 18.98 & 17.15 & 4.19 & 11.78 & 9.27  & 10.30 & 1.83 & 12.33 & 61.03 & 31.17 & 34.78 & 14.66 & 12.40 \\
QueryOcc~\cite{lilja2026queryocc} & 21.3 & \underline{55.0} & 7.3 & 6.8 & \underline{26.5} & 20.9 & 4.8 & 10.9 & \textbf{15.0} & \underline{13.2} & 3.7 & 17.3 & \textbf{69.2} & 34.5 & 38.4 & 25.2 & \underline{25.7}\\
FreeOcc (+ STCOcc~\cite{liao2025stcocc})~\cite{caunes2026freeocc} & 22.8 & 43.8 & - & - & - & - & - & - & - & - & - & - & - & - & - & - & -\\
ShelfOcc (+ STCOcc~\cite{liao2025stcocc})~\cite{boeder2025shelfocc}    & 22.87 & \textbf{56.14} & 13.98 & \underline{11.36} & 25.27 & 25.80 & 7.25 & \textbf{16.61} & 12.91 & \textbf{13.42} & 5.37 & 17.15 & \underline{68.01} & 34.66 & \textbf{42.73} & 25.63 & 22.89 \\
LetOccVote-L~\cite{zhang2026letoccvote} & \underline{23.11} & 53.27 & 8.36 & \textbf{13.12} & 22.02 & \underline{29.49} & \underline{12.76} & \underline{16.12} & 13.52 & 9.48 & \underline{7.38} & \underline{18.66} & 66.52 & \underline{36.98} & \underline{40.00} & \textbf{26.02} & \textbf{26.20}\\
\midrule
\rowcolor{bannergray}
\approach (Ours) (+ STCOcc~\cite{liao2025stcocc}) & \textbf{24.86} & 51.13 & \underline{21.46} & 10.78 & \textbf{28.70} & \textbf{30.12} & \textbf{15.91} & 14.48 & 14.25 & 10.55 & \textbf{9.77} & \textbf{24.24} & 63.73 & \textbf{40.83} & 38.29 & \underline{25.71} & 24.14\\
\bottomrule
\end{tabular}
\end{table*}

\begin{table*}[t]
  \centering
  \setlength{\tabcolsep}{5pt}
  \scriptsize
  \caption{Track-level accuracy of the motion segmentation strategies on the nuScenes validation set, restricted to the Occ3D range.
  We use all annotated tracks that are visible and are segmented by a SAM3.1 track in at least one view, which are identical for all methods.
  We consider a track ground-truth moving if its nuScenes attribute determines it as moving at any keyframe. %
  \textit{Leaked moving tracks} ($1-R_\text{mov}$) measures the share of moving tracks that get accumulated into the static map and produce motion trails.
  \textit{Moving share of accumulated tracks} ($1-P_\text{stat}$) represents the share of accumulated tracks that are actually moving.
  \textit{Recovered static tracks} ($R_\text{stat}$) is the share of static tracks of potentially dynamic classes that are accumulated.
  Since a track counts as recovered only if none of its observations is marked as dynamic, $R_\text{stat}$ is a lower bound.
  The majority variants instead decide per track based on a majority of its observations, meaning a static track is recovered if fewer than half of its observations are marked as dynamic.
  Conversely, a moving track pollutes the static map if at least half of its observations are marked static.
  Both must be read together, since a segmentation that marks nothing as dynamic reaches $R^\text{maj}_\text{stat}=100$.}
  \label{tab:supp_motion_seg}
  \begin{tabular}{lccccc}
    \toprule
    Method & Leaked moving & Moving share of & Recovered static & Leaked moving tracks & Recovered static tracks \\
           & tracks ($1-R_\text{mov}$) $\downarrow$ & accumulated tracks ($1-P_\text{stat}$) $\downarrow$ & tracks ($R_\text{stat}$) $\uparrow$ & by majority ($1-R^\text{maj}_\text{mov}$) $\downarrow$ & by majority ($R^\text{maj}_\text{stat}$) $\uparrow$ \\
    \midrule
    Class-based dynamics proxy~\cite{boeder2025shelfocc, caunes2026freeocc} & 0.0  & --   & 0.0  & 0.0  & 0.0 \\
    \midrule
    TT-Occ-Camera~\cite{zhang2026test}                                              & 14.6 & 33.7 & 20.7 & 60.6 & \textbf{72.5} \\
    \midrule
    \rowcolor{bannergray}
    LTMS (Ours + MapAnything~\cite{keetha2026mapanything}) & \textbf{4.4} & 10.9 & 26.0 & \textbf{10.6} & 51.9 \\
    \rowcolor{bannergray}
    LTMS (Ours + Pi3X~\cite{wang2026pi}) & 4.9 & \textbf{10.4} & \textbf{30.4} & 12.1 & 56.2\\
    \bottomrule
  \end{tabular}
  \vspace{-0.2cm}
\end{table*}

\subsection{Per-Class Performance}
{\parskip=2pt
\noindent\textbf{4D Occupancy Forecasting.}
We evaluate the per-class performance of \approach on Occ3D-nuScenes in \Cref{tab:per_class_forecasting}.
Compared to SparseWorld-TC trained fully supervised with the 3D Occ3D labels, our model shows only a slight decrease in accuracy for large stuff classes, such as driveable surface and sidewalk.
For the large vehicle classes bus, car, and truck, which we specifically target with our proposed components, \approach retains about two-thirds of IoU.
The largest difference lies in classes with small objects (bicycle, motorcycle, and traffic cone), while both models show inferior results for pedestrians.
We attribute the difference to imperfections in the depth predictions of our pseudo-labels, which seem to propagate and more negatively influence the results for the underlying sparse model, compared to the voxel-based STCOcc (see \Cref{tab:occ3d_results}).
Since we exclude \textit{others} and \textit{other\_flat} from the prompts, their values are 0, and we additionally report mIoU* that excludes these two classes from the overall average.
Using SparseWorld-TC-Small* with DINOv3 initialization improves the results mainly for static and small classes, while performing slightly worse for large vehicle classes.
For the class-based results for RayIoU in \Cref{tab:per_class_forecasting_rayiou}, we observe an overall similar pattern, although the stuff classes score worse relative to the ground-truth training.
A reason for this is that RayIoU penalizes incorrect depth predictions of flat regions more than mIoU, and our approach does not rely on LiDAR depth.
In contrast, RayIoU for our targeted large vehicle classes shows a similar gap to full supervision as observed for mIoU.
}

{\parskip=2pt
\noindent\textbf{3D Occupancy Prediction.}
We provide per-class results for 3D occupancy estimation of STCOcc~\cite{liao2025stcocc} trained with our pseudo-labels in \Cref{tab:occ3d_results}.
Since we do not discard information about stationary objects from our pseudo-labels, our model improves especially on potentially moving classes with high proportions of stationary instances, such as bus, car, and truck, compared to previous state-of-the-art methods.
}

\subsection{Evaluation of Motion Masks}
In \Cref{tab:supp_motion_seg}, we evaluate the track-level accuracy of our proposed LTMS compared to a class-based dynamics proxy and TT-Occ-Camera~\cite{zhang2026test}.
Since the former excludes all potentially dynamic object classes from temporal point cloud accumulation, it does not leak any moving tracks but also does not recover any static object tracks.
Our LTMS, with either MapAnything~\cite{keetha2026mapanything} or Pi3X~\cite{wang2026pi}, leaks considerably fewer moving tracks into the static map than TT-Occ-Camera.
Moreover, far fewer accumulated tracks based on LTMS are actually moving, while it still achieves higher recall for static objects.
The gap in pollution widens when each track is judged by the majority of its observations.
In this case, TT-Occ leaks most of the moving tracks ($60.6\%$) into the static map, whereas LTMS keeps this rate at $12.1\%$ with Pi3X.
Under this rule, TT-Occ recovers more static tracks, but only because it marks few observations as dynamic, leading to a large number of leaked moving tracks.
These results are in line with our design goal of LTMS, which is to conservatively accumulate points from truly static objects, while strictly avoiding object trails in the static point cloud.

\FloatBarrier

\section{Additional Qualitative Results}\label{sec:supp_qualitative}

\begin{figure*}[t]
    \centering
    {\includegraphics[width=0.975\linewidth]{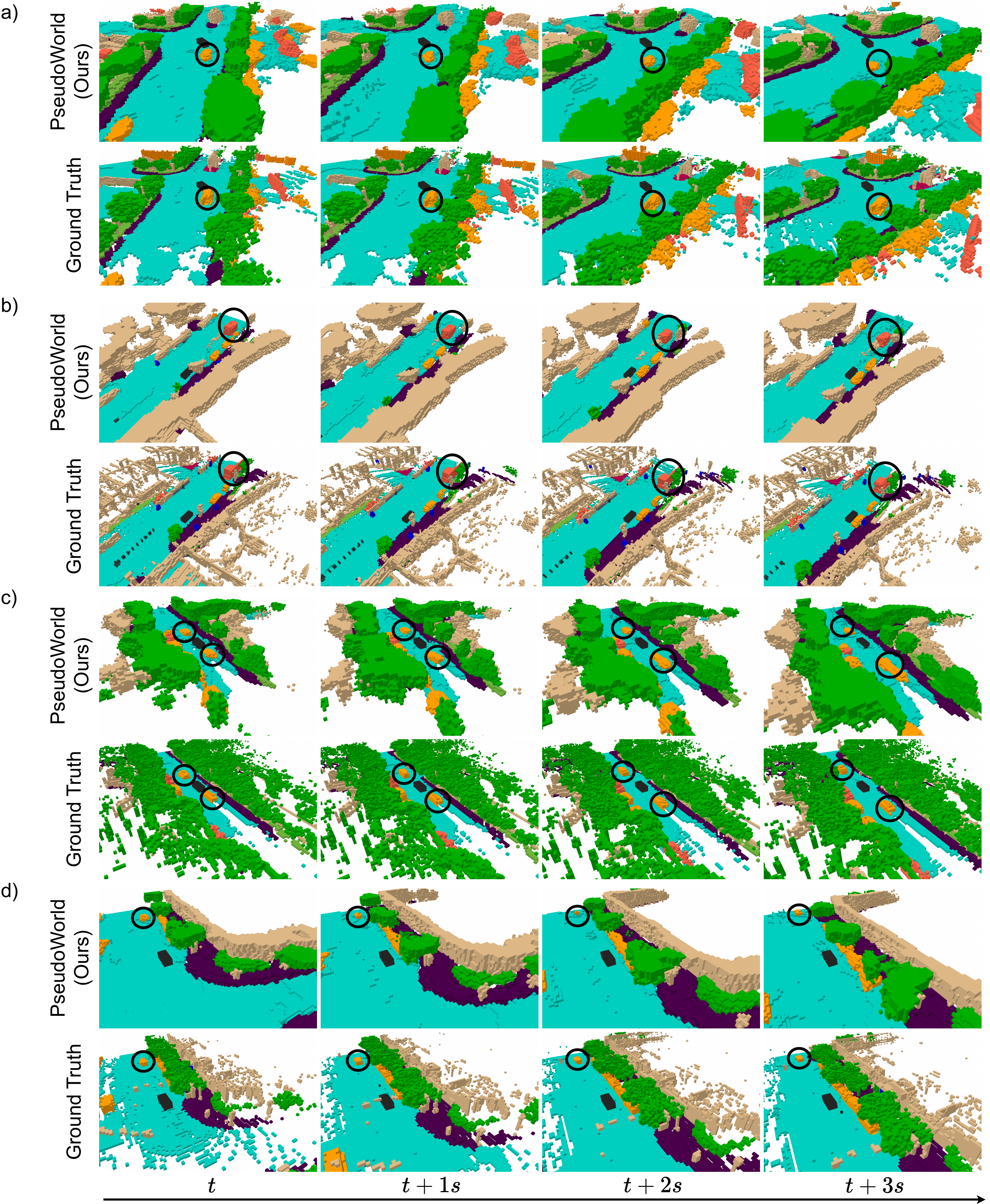}}
    \caption{Qualitative results on the Occ3D-nuScenes dataset. In a), \approach correctly predicts the motion of the following car. In b), \approach detects and forecasts the stopped truck that the ego-vehicle approaches correctly. In c) and d), our approach demonstrates that it is able to forecast the movement of cars driving ahead of and behind the ego-vehicle.}
    \label{fig:qualitative_supp}
    \vspace{-0.2cm}
\end{figure*}

In \Cref{fig:qualitative_supp}, we provide additional qualitative results of \approach in comparison with the Occ3D-nuScenes ground truth labels.
Our approach demonstrates that it can recover parked objects well and also estimate the motion of moving vehicles correctly.

\end{document}